\documentclass[sigconf,nonacm]{acmart}

\setcopyright{none}
\renewcommand\footnotetextcopyrightpermission[1]{}

\usepackage{graphicx}
\usepackage{subcaption}
\usepackage{enumitem}
\usepackage{booktabs}
\usepackage{mathtools}
\usepackage{amsthm}
\usepackage{interval}
\usepackage{multirow}
\usepackage{makecell}
\usepackage[capitalize,noabbrev]{cleveref}

\theoremstyle{plain}

\theoremstyle{definition}

\theoremstyle{remark}

\acmConference{}{}{}
\acmDOI{}
\acmISBN{}

\title{Toward Reliable LLM Evaluation: Process-Aware Measurement and Auditable Collaboration}

\author{Yifan Yang}
\authornote{Equal contribution.}
\affiliation{\institution{The Hong Kong Polytechnic University}\country{Hong Kong SAR}}
\author{Xijun Xiang}
\authornotemark[1]
\affiliation{\institution{The Hong Kong Polytechnic University}\country{Hong Kong SAR}}
\author{Yuanyi Wang}
\affiliation{\institution{The Hong Kong Polytechnic University}\country{Hong Kong SAR}}
\author{Jinjia Li}
\affiliation{\institution{InfiX.ai}}
\author{Puhao Zheng}
\affiliation{\institution{The Hong Kong Polytechnic University}\country{Hong Kong SAR}}
\author{Ningcong Gao}
\affiliation{\institution{The Hong Kong Polytechnic University}\country{Hong Kong SAR}}
\author{Zhijie Sang}
\affiliation{\institution{InfiX.ai}}
\author{Jianmin Wu}
\affiliation{\institution{InfiX.ai}}
\author{Ming Li}
\authornote{Corresponding authors.}
\affiliation{\institution{The Hong Kong Polytechnic University}\country{Hong Kong SAR}}
\email{ming.li@polyu.edu.hk}
\author{Hongxia Yang}
\authornotemark[2]
\affiliation{\institution{The Hong Kong Polytechnic University}\country{Hong Kong SAR}}
\email{hongxia.yang@polyu.edu.hk}
\renewcommand{\shortauthors}{Yang et al.}
\hypersetup{pdfauthor={Yifan Yang, Xijun Xiang, Yuanyi Wang, Jinjia Li, Puhao Zheng, Ningcong Gao, Zhijie Sang, Jianmin Wu, Ming Li, Hongxia Yang}}

\makeatletter
\gdef\@currentauthors{}
\renewcommand{\@mkauthors}{%
  \global\setbox\mktitle@bx=\vbox{%
    \unvbox\mktitle@bx\par\medskip
    \centering
    {\normalsize
      Yifan Yang\textsuperscript{1,*}\quad
      Xijun Xiang\textsuperscript{1,*}\quad
      Yuanyi Wang\textsuperscript{1}\quad
      Jinjia Li\textsuperscript{2}\quad
      Puhao Zheng\textsuperscript{1}\par
      Ningcong Gao\textsuperscript{1}\quad
      Zhijie Sang\textsuperscript{2}\quad
      Jianmin Wu\textsuperscript{2}\quad
      Ming Li\textsuperscript{1,\textdagger}\quad
      Hongxia Yang\textsuperscript{1,\textdagger}\par}
    \vskip3pt
    {\small
      \textsuperscript{1}The Hong Kong Polytechnic University, Hong Kong SAR\quad
      \textsuperscript{2}InfiX.ai\par
      \href{mailto:ming.li@polyu.edu.hk}{ming.li@polyu.edu.hk}\quad
      \href{mailto:hongxia.yang@polyu.edu.hk}{hongxia.yang@polyu.edu.hk}\par}
    \vskip8pt}}
\makeatother

\begin{document}

\begin{abstract}
Black-box benchmark evaluation is widely used to compare LLMs, yet reported scores alone provide an incomplete picture of outcome reliability, reasoning stability, and the trustworthiness of evaluation procedures.
At the outcome level, our analysis of 50 configuration-matched rounds reveals ranking inversion probabilities reaching 42\%.
Existing work often assesses outcome reliability through repeated evaluation, requiring multiple executions to characterize score variability and estimation uncertainty. 
However, even stable aggregate scores provide limited evidence about stepwise reasoning support or the trustworthiness of evaluation procedures, leaving process reliability and procedural integrity insufficiently addressed.
We introduce a process-level evaluation method Process-Aware Likelihood Support (PALS), which extracts stepwise predictive-support evidence in a single scoring round.
Interpreted through sequence-level conditional cross-mutual information, PALS quantifies the incremental predictive contribution of a preceding reasoning step by comparing the same target with that step retained or removed.
Complementing these process diagnostics, we develop CoEvalChain (A Collaborative Evaluation Blockchain), a framework for collaborative LLM evaluation that supports outcome reliability and procedural integrity through repeated assessment, robust aggregation, blockchain-backed commitments, and auditable reporting.
Controlled experiments show that PALS responds to disrupted reasoning dependencies and captures support variability across repeatedly sampled next steps.
Across five open-weight models and five benchmarks, the collaborative protocol yields lower round-level score dispersion than centralized evaluation in 22 of 25 settings.
Single-round PALS measurements on shared references and model-generated traces show negative associations with repeated-score variability, providing complementary evidence beyond outcome scores.
Our integrity analysis further clarifies the scope of CoEvalChain's procedural safeguards and the trust assumptions underlying reliable collaborative aggregation.
To support reproducible evaluation, we published the PALS DAG-construction code and reference CoT data publicly available at https://anonymous.4open.science/r/pals-8F67/. And we will release the CoEvalChain web platform later.
\end{abstract}



\makeatletter
\let\coeval@saved@specialsection\@specialsection
\pretocmd{\@specialsection}{%
  \patchcmd{\section}
    {-.75\baselineskip \@plus -2\p@ \@minus -.2\p@}
    {-3pt}{}{}%
  \patchcmd{\section}{.25\baselineskip}{1pt}{}{}%
}{}{}
\makeatother
\raggedbottom
\maketitle
\makeatletter
\let\@specialsection\coeval@saved@specialsection
\makeatother
\flushbottom

\begingroup
\raggedbottom
\begingroup

\makeatletter
\patchcmd{\section}
  {-.75\baselineskip \@plus -2\p@ \@minus -.2\p@}
  {-3pt \@plus -.5pt \@minus -.5pt}{}{}
\patchcmd{\section}{.25\baselineskip}{1pt}{}{}
\patchcmd{\subsection}
  {-.75\baselineskip \@plus -2\p@ \@minus -.2\p@}
  {-2pt \@plus -.5pt \@minus -.5pt}{}{}
\patchcmd{\subsection}{.25\baselineskip}{1pt}{}{}
\patchcmd{\subsubsection}
  {-.5\baselineskip \@plus -2\p@ \@minus -.2\p@}
  {-1pt \@plus -.5pt \@minus -.5pt}{}{}
\patchcmd{\paragraph}
  {-.5\baselineskip \@plus -2\p@ \@minus -.2\p@}
  {-1pt \@plus -.5pt \@minus -.5pt}{}{}
\makeatother

\setlength{\parskip}{-1pt}
\setlength{\abovedisplayskip}{1.5pt plus 0.5pt minus 0.5pt}
\setlength{\belowdisplayskip}{1.5pt plus 0.5pt minus 0.5pt}
\setlength{\abovedisplayshortskip}{0pt plus 1pt}
\setlength{\belowdisplayshortskip}{1.5pt plus 0.5pt minus 0.5pt}

\setlength{\textfloatsep}{8pt plus 2pt minus 2pt}
\setlength{\dbltextfloatsep}{8pt plus 2pt minus 2pt}
\setlength{\floatsep}{6pt plus 2pt minus 2pt}
\setlength{\dblfloatsep}{6pt plus 2pt minus 2pt}
\setlength{\intextsep}{8pt plus 2pt minus 2pt}
\captionsetup{font=small,skip=4pt}

\begin{figure}[!t]
    \centering
    \includegraphics[width=0.88\linewidth]{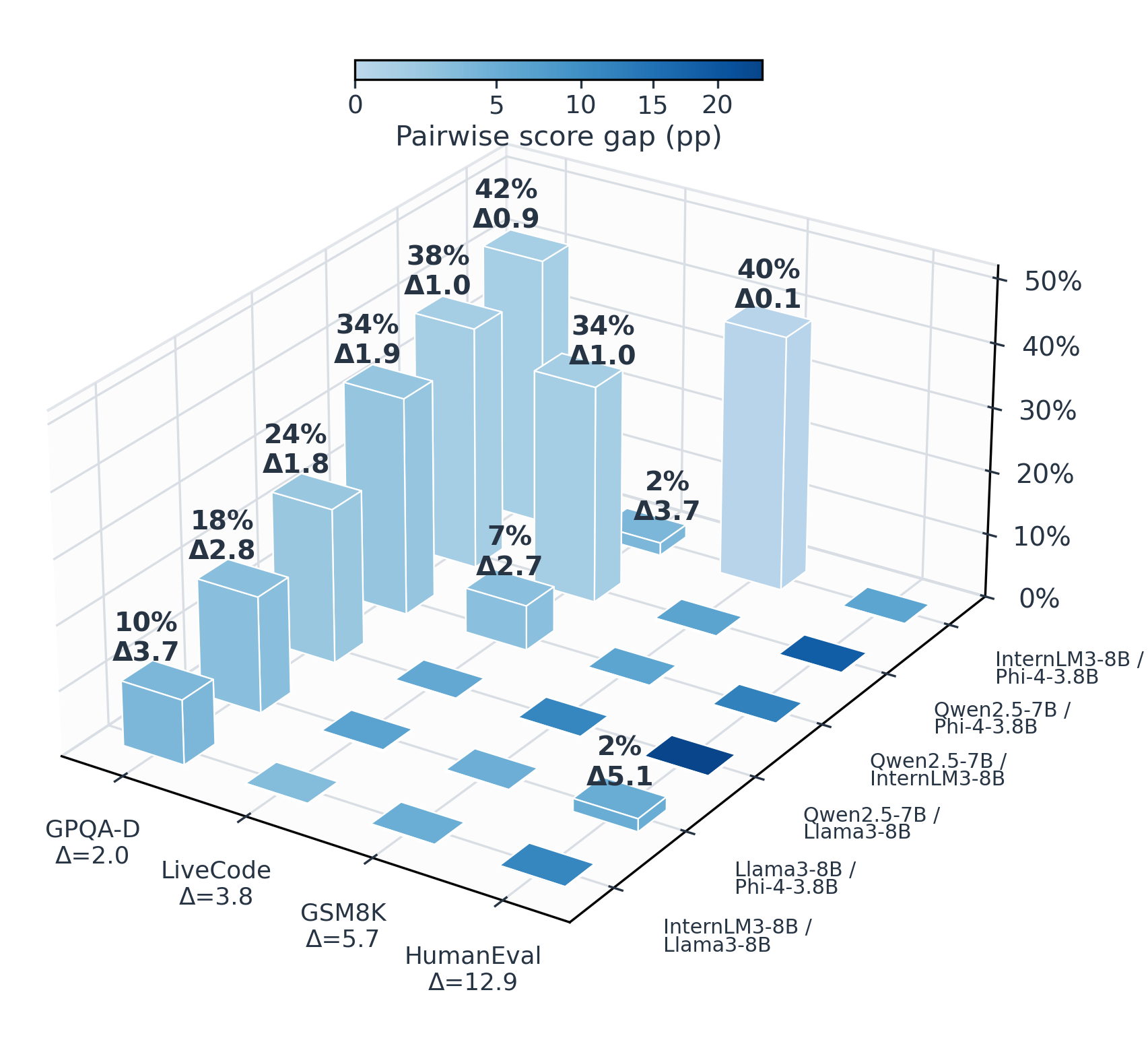}
    \caption{\textbf{RIP and score separation.} Bar height shows RIP; color and $\Delta$ indicate pairwise score gaps in percentage points. Benchmark labels report average gaps across model pairs.}
    \Description{A three-dimensional bar chart comparing six pairs of similarly sized language models across four benchmarks. Bar height represents ranking inversion probability, and blue intensity represents the pairwise score gap for each model-pair and benchmark combination.}
    \label{fig:rip_heatmap}
\end{figure}

\section{Introduction}
\label{sec:introduction}

Benchmark scores guide large language model (LLM) selection and track research progress \citep{refer04,refer23}, yet a single reported score can obscure the stochastic nature of evaluation. Existing approaches strengthen evaluation through task-specific benchmarks~\citep{refer18}, uncertainty estimation~\citep{refer30}, analyses of generation inconsistency~\citep{refer24}, and human preference comparisons~\citep{refer27}. These efforts improve the evidence available for model comparison, but decoding randomness and differences in inference settings, hardware, and numerical execution can still change model outputs. When this variation is comparable to the gap between models, leaderboard positions may reflect individual executions rather than reproducible capability \citep{refer30,yuan2025numericalnondeterminism}.\looseness=-1

We frame reliable LLM evaluation along three complementary dimensions: \emph{outcome reliability}, concerning score variation, estimation uncertainty, and ranking stability; \emph{process reliability}, concerning the consistency of observable support relationships within reasoning; and \emph{procedural integrity}, concerning the traceability and auditability of evaluation procedures. Together, these dimensions connect reported performance with the reasoning being assessed and the procedures used to assess it.\looseness=-1

To examine outcome reliability, we analyze ranking fragility through 50 configuration-matched rounds for four similarly sized models on GSM8K \citep{gsm8k}, HumanEval \citep{humaneval}, GPQA-Diamond \citep{refer31}, and LiveCodeBench \citep{livecodebench}. We report \emph{ranking inversion probability} (RIP): the probability that a paired single run reverses the ordering by mean scores, with ties assigned half weight in Figure~\ref{fig:rip_heatmap}. The figure reveals two implications for reliable model comparison. First, benchmark-level discrimination does not guarantee reliable pairwise ordering: GSM8K has an average score gap of 5.7 percentage points, yet InternLM3-8B and Phi-4-mini differ by only 0.1 points and exhibit 40\% RIP. A benchmark can therefore separate models well on average while leaving close competitors difficult to distinguish in a single run. Second, RIP ranges from 10\% to 42\% across model pairs within GPQA-Diamond, showing that ranking stability is a property of a particular comparison under an evaluation protocol. Neither a model nor a benchmark should be labeled globally ``stable'' or ``unstable'' from such rankings, reliability depends on which models are compared, on which task, and how their score separation relates to paired execution variability. Repeated evaluation should therefore establish how consistently an observed advantage persists, alongside reporting its magnitude. However, a sufficiently large margin can preserve the ordering even when individual outputs vary. Since stable rankings alone do not reflect stable reasoning, it motivates complementary assessment of the generation process.\looseness=-1

Equal accuracy, or even identical answers, can arise from different intermediate reasoning steps. Existing approaches such as SPIRIT~\citep{cui2025spirit} and RISE~\citep{sengupta2026rise} do not directly characterize both the signed support a preceding step provides for a fixed next step and its variability across generations. To assess process reliability beyond these outcomes, we propose \textbf{Process-Aware Likelihood Support (PALS)}, using externally prepared reference reasoning traces to provide controlled reasoning contexts. PALS holds the target step and earlier prefix fixed and measures the signed gain in the target's average log-likelihood from retaining its immediately preceding step. A single scoring round characterizes support levels on a fixed trace, while repeated next-step sampling from the same prefix reveals support variability. These diagnostics complement outcome statistics without, by themselves, establishing logical correctness or identifying the model's internal causal mechanisms.\looseness=-1

While PALS complements outcome scores with process-level evidence, trustworthy collaborative evaluation also requires procedural integrity in how contributors execute assessments and report their results. Repetition alone does not address fabricated reports, copied submissions, or coordinated manipulation \citep{zhang2019securecrowdsourcing,cai2020truthoracle}. We therefore design \textbf{CoEvalChain}, a collaborative framework that combines repeated assessment and robust aggregation with verifiable coordination and reporting. Shared task specifications and recorded execution configurations make reports interpretable, while blockchain-based commitments, commit--reveal submissions, and auditable incentives support accountability throughout the evaluation workflow \citep{zhang2019securecrowdsourcing,cai2020truthoracle,canidio2022commitreveal}. CoEvalChain thus addresses both the precision of performance estimates and the integrity of the procedures used to obtain them; its audit mechanisms do not independently guarantee honest off-chain computation. PALS supplies complementary process diagnostics without altering the canonical benchmark score. Figure~\ref{fig:pals-coevalchain-overview} presents the combined workflow.\looseness=-1

Together, PALS and CoEvalChain connect these three dimensions of reliable evaluation. PALS provides stepwise support diagnostics for \emph{process reliability}, while CoEvalChain supports \emph{outcome reliability} through repeated assessment and aggregation, and \emph{procedural integrity} through traceable coordination and auditable reporting. For each model–benchmark setting, we compare average support from one PALS scoring round with score dispersion across repeated evaluations. This comparison examines the association between stepwise predictive support and outcome variability.

Our main contributions are as follows:

\begin{itemize}[topsep=2pt,itemsep=0pt,parsep=0pt,partopsep=0pt,leftmargin=1.1em,labelsep=.35em]
  \item We propose PALS, a new process-level method for assessing LLM reasoning stability. Controlled experiments show that PALS reports lower average support when reasoning dependencies are disrupted and captures support variation across repeated generations.
  \item We design and implement CoEvalChain, a blockchain-backed framework for repeated collaborative LLM evaluation. Experiments show lower round-level score dispersion in 22 of 25 model–benchmark settings.
  \item We integrate PALS and CoEvalChain into a unified framework for stable and trustworthy LLM evaluation. This integration jointly addresses process reliability, outcome reliability, and procedural integrity by connecting reasoning diagnostics with repeated, auditable performance assessment. Experiments support this connection: higher PALS support on reference traces is associated with lower CoEvalChain score dispersion (Spearman $\rho=-0.357$).
\end{itemize}

\begin{figure*}[!t]
    \centering
    \includegraphics[width=\textwidth]{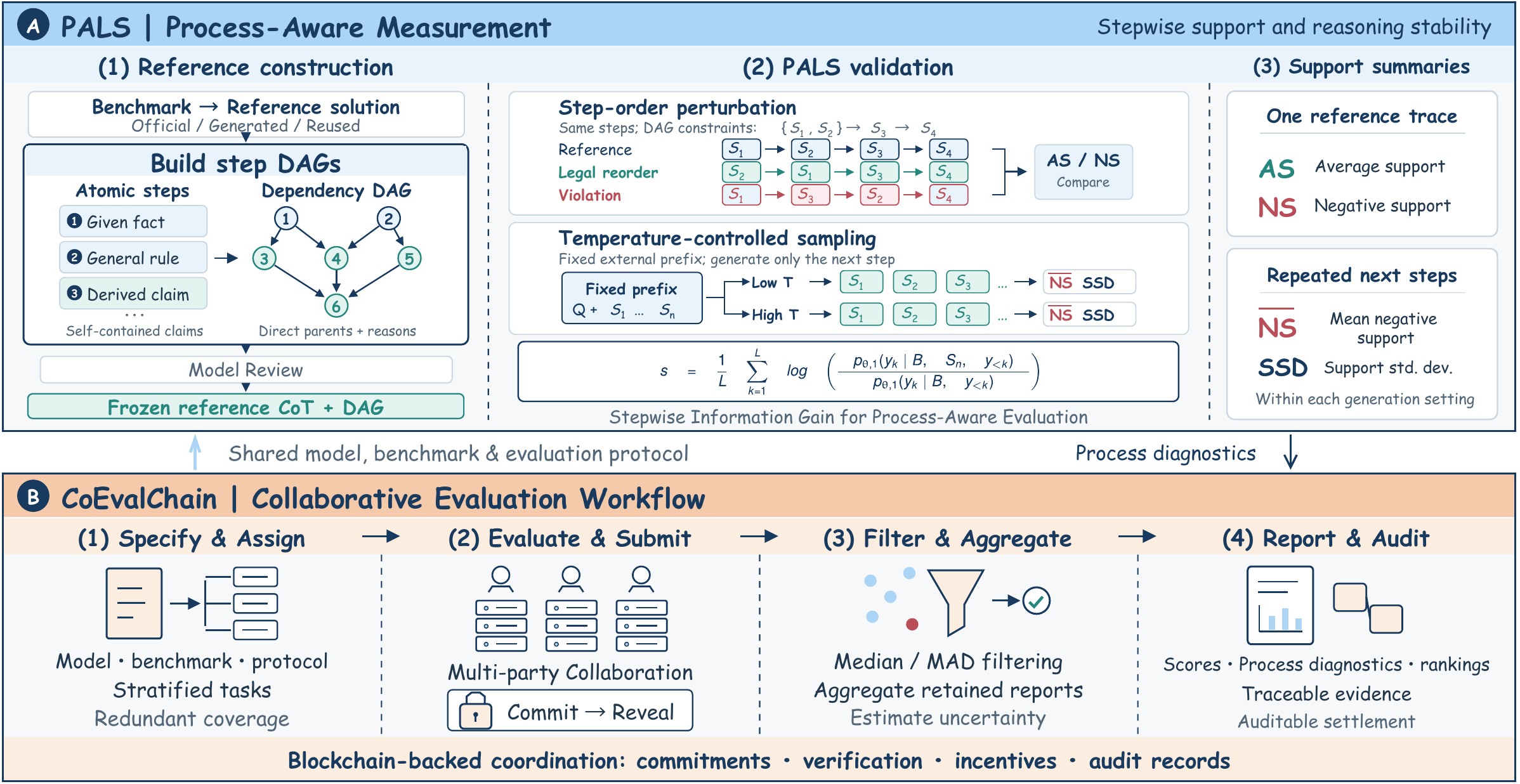}
    \caption{Overview of process-aware measurement and collaborative evaluation. The upper layer constructs reference reasoning DAGs, validates PALS through controlled step reordering and next-step sampling, and summarizes stepwise support. The lower layer coordinates task assignment, evaluation, aggregation, and auditable reporting through CoEvalChain.}
    \Description{A two-layer overview. The upper PALS layer contains reference construction, validation by step-order perturbation and temperature-controlled sampling, and support summaries. The lower CoEvalChain layer contains task specification and assignment, evaluation and submission, filtering and aggregation, and reporting and audit, supported by blockchain-based coordination.}
    \label{fig:pals-coevalchain-overview}
\end{figure*}

\section{Problem Formulation and Reliability Background}
\label{sec:reliability}

\subsection{Evaluation Setting}
\label{subsec:eval-setting}
Let $\theta$ denote a model, $\mathcal B$ a benchmark, $v$ an evaluator, and $r$ a repeated evaluation round. Configuration $c_{v,r}$ specifies prompts, decoding parameters, and the execution environment. For question $Q_q$, the output and benchmark score are
\begin{equation}
\label{eq:eval-observation}
\begin{aligned}
Y_{q,v,r}&\sim p_\theta(\cdot\mid Q_q;c_{v,r}),\\
s_{v,r}&=F_{\mathcal B}(\{Y_{q,v,r}\}_{q\in\mathcal B}),
\end{aligned}
\end{equation}
where $F_{\mathcal B}$ is the prescribed scoring and aggregation rule. The submitted score $\widetilde s_{v,r}$ may differ from the executed score $s_{v,r}$. An evaluator is a person or organization executing and reporting an assessment, not necessarily an LLM judge.

\subsection{Sources of Unreliability}
\label{subsec:reliability-sources}
\textbf{Execution and configuration.}
Sampling randomness, protocol differences, and numerical execution affect evaluation results. Prompt formatting and option order can change performance~\citep{sclar2024formatspread,pezeshkpour2024optionorder}, while dataset coverage and scoring implementations affect comparability~\citep{biderman2024lessons}. Hardware and numerical differences can yield divergent outputs even under greedy decoding~\citep{yuan2025numericalnondeterminism}. These factors affect both within-configuration variability and mean scores across configurations, so pooled results require a specified configuration distribution for meaningful interpretation.\looseness=-1

\textbf{Participants and reporting.}
Execution mistakes and reporting errors can compromise results, while participants may selectively report favorable runs, fabricate scores, or coordinate submissions. These risks persist under standardized configurations: coordinated reports can agree yet remain biased, so low dispersion alone does not establish trustworthy evaluation.

\subsection{Reliability Objectives}
\label{subsec:reliability-dimensions}
\label{subsec:pals-motivation}
\label{subsec:collaboration-motivation}
\textbf{Outcome reliability.}
For fixed $\theta$ and $\mathcal B$, let $\pi$ specify the distribution of evaluation configurations. Expectations and variances below account for both configuration variation and generation randomness. Define
\begin{equation}
\label{eq:outcome-reliability}
\begin{gathered}
\mu_\pi=\mathbb E_\pi[s_{v,r}],\qquad
\sigma_\pi^2=\operatorname{Var}_\pi(s_{v,r}),\\[2pt]
\mathbb E\!\left[(\widehat\mu-\mu_\pi)^2\right]
=\operatorname{Var}(\widehat\mu)
+\operatorname{Bias}(\widehat\mu)^2.
\end{gathered}
\end{equation}
Here, $\mu_\pi$ and $\sigma_\pi^2$ denote the mean and variance of executed scores, while $\widehat\mu$ estimates $\mu_\pi$ from repeated evaluations. The last equality decomposes the estimator's mean squared error into its variance and squared bias. Repetition can improve the precision of this estimate, but does not by itself eliminate bias or stabilize individual generations~\citep{refer30}.

\textbf{Process reliability.}
For a reference trace $\tau$ with $m$ scored targets $Y_i$ and contexts $C_i$, represent its support profile as
\begin{equation}
\label{eq:process-reliability}
\mathbf u_\theta(Q,\tau)=(u_1,\ldots,u_m),\qquad
u_i=\phi_\theta(C_i,Y_i).
\end{equation}
PALS (Section~\ref{sec:pals}) instantiates the support function $\phi_\theta$. A single scoring round describes support levels on a fixed trace; repeatedly sampling and scoring next steps from the same fixed context reveals support variability.

\textbf{Procedural integrity.}
For a fixed set of reports and aggregation rule $\mathcal A$, reporting deviations affect the aggregate through
\begin{equation}
\label{eq:reporting-integrity}
\begin{aligned}
\widetilde s_{v,r}&=s_{v,r}+\varepsilon_{v,r},\\
\Delta_{\mathrm{report}}
&=\mathcal A(\{\widetilde s_{v,r}\})-\mathcal A(\{s_{v,r}\}).
\end{aligned}
\end{equation}
For evaluator $v$ in round $r$, $\varepsilon_{v,r}$ equals the reported score minus the score obtained from execution. $\Delta_{\mathrm{report}}$ measures the resulting change in the aggregate when the aggregation rule is fixed. CoEvalChain (Section~\ref{sec:coevalchain}) additionally records task coverage, configurations, and commitments for audit, commitment consistency alone does not establish $\varepsilon_{v,r}=0$.

\begin{table}[t]
\centering
\captionsetup{font=small}
\caption{E1: Support under step reordering. Entries report AS / NS in nats per token; baseline denotes the branch-contiguous order.}
\label{tab:pals-e1}
\fontsize{7.5}{8.5}\selectfont
\setlength{\tabcolsep}{2pt}
\begin{tabular*}{\columnwidth}{@{\extracolsep{\fill}}lrrr@{}}
\toprule
Model & \makecell[r]{Baseline\\AS / NS} & \makecell[r]{Legal\\AS / NS} & \makecell[r]{Disruptive\\AS / NS} \\
\midrule
\multicolumn{4}{@{}l}{\textit{GPQA-Diamond}} \\
Qwen2.5-7B & $0.2715 \,/\, 0.0397$ & $0.3076 \,/\, 0.0345$ & $0.2268 \,/\, 0.0491$ \\
Phi-4-mini & $0.2369 \,/\, 0.0261$ & $0.2582 \,/\, 0.0236$ & $0.1996 \,/\, 0.0316$ \\
Qwen3-14B & $0.4139 \,/\, 0.0248$ & $0.4324 \,/\, 0.0217$ & $0.3684 \,/\, 0.0324$ \\
Llama-3-8B & $0.2391 \,/\, 0.0234$ & $0.2587 \,/\, 0.0219$ & $0.1969 \,/\, 0.0297$ \\
InternLM3-8B & $0.2155 \,/\, 0.0208$ & $0.2162 \,/\, 0.0173$ & $0.1742 \,/\, 0.0285$ \\
\addlinespace[2pt]
\multicolumn{4}{@{}l}{\textit{HumanEval}} \\
Qwen2.5-7B & $0.3272 \,/\, 0.0232$ & $0.3553 \,/\, 0.0170$ & $0.3095 \,/\, 0.0261$ \\
Phi-4-mini & $0.2413 \,/\, 0.0138$ & $0.2670 \,/\, 0.0098$ & $0.2112 \,/\, 0.0222$ \\
Qwen3-14B & $0.4833 \,/\, 0.0162$ & $0.4753 \,/\, 0.0170$ & $0.4548 \,/\, 0.0239$ \\
Llama-3-8B & $0.2608 \,/\, 0.0123$ & $0.2716 \,/\, 0.0105$ & $0.2397 \,/\, 0.0196$ \\
InternLM3-8B & $0.2254 \,/\, 0.0147$ & $0.2391 \,/\, 0.0096$ & $0.1985 \,/\, 0.0231$ \\
\addlinespace[2pt]
\multicolumn{4}{@{}l}{\textit{LiveCodeBench v6}} \\
Qwen2.5-7B & $0.2697 \,/\, 0.0102$ & $0.2711 \,/\, 0.0164$ & $0.2374 \,/\, 0.0176$ \\
Phi-4-mini & $0.2238 \,/\, 0.0104$ & $0.2150 \,/\, 0.0127$ & $0.1858 \,/\, 0.0219$ \\
Qwen3-14B & $0.4344 \,/\, 0.0048$ & $0.4309 \,/\, 0.0105$ & $0.3826 \,/\, 0.0155$ \\
Llama-3-8B & $0.2107 \,/\, 0.0072$ & $0.2145 \,/\, 0.0101$ & $0.1871 \,/\, 0.0173$ \\
InternLM3-8B & $0.1905 \,/\, 0.0082$ & $0.1906 \,/\, 0.0131$ & $0.1550 \,/\, 0.0153$ \\
\addlinespace[2pt]
\multicolumn{4}{@{}l}{\textit{GSM8K}} \\
Qwen2.5-7B & $0.4248 \,/\, 0.0425$ & $0.4418 \,/\, 0.0350$ & $0.1595 \,/\, 0.1628$ \\
Phi-4-mini & $0.3905 \,/\, 0.0481$ & $0.3916 \,/\, 0.0308$ & $0.1375 \,/\, 0.1572$ \\
Qwen3-14B & $0.7355 \,/\, 0.0311$ & $0.7334 \,/\, 0.0215$ & $0.4020 \,/\, 0.1561$ \\
Llama-3-8B & $0.3088 \,/\, 0.0336$ & $0.3522 \,/\, 0.0213$ & $0.1058 \,/\, 0.1362$ \\
InternLM3-8B & $0.2774 \,/\, 0.0191$ & $0.2709 \,/\, 0.0159$ & $0.0953 \,/\, 0.0999$ \\
\addlinespace[2pt]
\multicolumn{4}{@{}l}{\textit{MMLU}} \\
Qwen2.5-7B & $0.4529 \,/\, 0.0963$ & $0.4781 \,/\, 0.0802$ & $0.0941 \,/\, 0.2198$ \\
Phi-4-mini & $0.3927 \,/\, 0.0465$ & $0.3833 \,/\, 0.0410$ & $0.1201 \,/\, 0.1224$ \\
Qwen3-14B & $0.8005 \,/\, 0.0496$ & $0.8444 \,/\, 0.0496$ & $0.4638 \,/\, 0.1428$ \\
Llama-3-8B & $0.3469 \,/\, 0.0455$ & $0.3628 \,/\, 0.0412$ & $0.1130 \,/\, 0.1238$ \\
InternLM3-8B & $0.2666 \,/\, 0.0415$ & $0.2747 \,/\, 0.0321$ & $0.0944 \,/\, 0.0857$ \\
\bottomrule
\end{tabular*}
\end{table}

\section{Measuring Reasoning-Process Stability with PALS}
\label{sec:pals}

\subsection{Measurement Objective and Setting}
\label{subsec:pals-setting}
PALS measures the incremental predictive support of a preceding reasoning step, providing a local, model-relative diagnostic. It does not assess logical correctness or all sources of process instability. Given a question $Q$, let $\tau=(S_1,\ldots,S_n)$ be an externally prepared reference reasoning trace, fixed before evaluation after model-based quality review and additional human verification where documented. A target is either a reference step or \emph{next-step continuation}: one reasoning step sampled by the tested model from a fixed reference prefix. PALS scores the same target with the immediately preceding step retained or removed. Repeatedly sampling next steps from the same prefix under fixed generation settings reveals variation in their predictive support.\looseness=-1

\subsection{PALS: Information-Theoretic Formulation and Stability Measurement}
\label{subsec:pals-support}
\label{subsec:pals-aggregation}
\label{subsec:pals-cxmi}
Conditional cross-mutual information (CXMI) was introduced to quantify context usage in context-aware machine translation~\citep{fernandes2021cxmi}. It compares predictive cross-entropies with and without additional context. Here, we use this contrast to measure the predictive support supplied by context $S$ for a target sequence $Y$, conditional on a retained background $B$. For an evaluation distribution $\rho(B,S,Y)$ and model $p_\theta$, define $H_{\rho,\theta}(Y\mid X)=\mathbb E_\rho[-\log p_\theta(Y\mid X)]$. Over complete target sequences with a consistent termination rule,
\begin{equation}
\label{eq:pals-cxmi}
\begin{aligned}
\mathrm{CXMI}_{\rho,\theta}(S\to Y\mid B)
&=H_{\rho,\theta}(Y\mid B)-H_{\rho,\theta}(Y\mid B,S)\\
&=\mathbb E_\rho\!\left[\log\frac{p_\theta(Y\mid B,S)}{p_\theta(Y\mid B)}\right].
\end{aligned}
\end{equation}
This gain is model-relative because both predictive distributions are evaluated under $p_\theta$.

PALS adapts this contrast to quantify the incremental predictive support of the immediately preceding reasoning step for a fixed target step.
For a target step $Y_i=(y_{i,1},\ldots,y_{i,L_i})$ at position $i\geq2$, define the full prefix and the counterpart obtained by deleting its last step:
\begin{equation}
\label{eq:pals-contexts}
C_i=(Q,S_1,\ldots,S_{i-1}),\quad
C_i^-=(Q,S_1,\ldots,S_{i-2}).
\end{equation}
Here $B=C_i^-$ and $S=S_{i-1}$; the target is either the reference step $S_i$ or a next-step continuation sampled from $C_i$. Normalizing the predictive gain by the target length gives the step support score
\begin{equation}
\label{eq:pals-step-support}
s_i=\frac{1}{L_i}\sum_{k=1}^{L_i}
\log\frac{p_\theta(y_{i,k}\mid C_i,y_{i,<k})}
{p_\theta(y_{i,k}\mid C_i^-,y_{i,<k})}.
\end{equation}
Thus, $\mathrm{CXMI}_{\rho,\theta}(S\to Y\mid B)=\mathbb E_\rho[L(Y)s(Y)]$, while $s$ measures the length-normalized gain for an individual reasoning step. Once $Y_i$ is obtained, its text is fixed while its token log-probabilities are evaluated under the full context $C_i$ and the deleted context $C_i^-$. Both evaluations use the same within-target token prefix $y_{i,<k}$ at each position; the target is not regenerated after deletion. Averaging the token log-probability differences using natural logarithms gives support in nats per token.

Positive $s_i$ means that the preceding step increases the target's likelihood, while negative $s_i$ indicates reduced likelihood. Near-zero support means little incremental predictive effect: retained context or model knowledge may already supply similar information. Equivalently, the score in Eq.~\eqref{eq:pals-step-support} is the difference between the negative log-likelihoods (NLLs) under the deleted and full contexts, with both NLLs averaged over the same target tokens:
\begin{equation}
\label{eq:pals-nll}
s_i=\mathrm{NLL}_\theta(Y_i\mid C_i^-)
-\mathrm{NLL}_\theta(Y_i\mid C_i).
\end{equation}
Whereas full-context NLL describes the target's absolute predictive difficulty, this contrast identifies the incremental support supplied by the preceding step.

The same contrast provides an interventional view of premise use in textual inference. With the model, background, and target fixed, PALS measures how removing the preceding step affects the model's prediction of the target~\citep{li2017erasure}. To examine whether this response follows reasoning dependencies, we organize reference reasoning into directed acyclic graphs (DAGs), whose nodes are atomic statements and edges denote direct premise dependencies. E1 in Section~\ref{subsec:pals-validation} compares dependency-preserving and dependency-disrupting reorderings using these annotations. Together, the score and structural controls probe causal sensitivity in textual inference: how preceding reasoning influences support for a subsequent step.

\begin{figure}[!t]
\centering
\includegraphics[width=0.88\columnwidth]{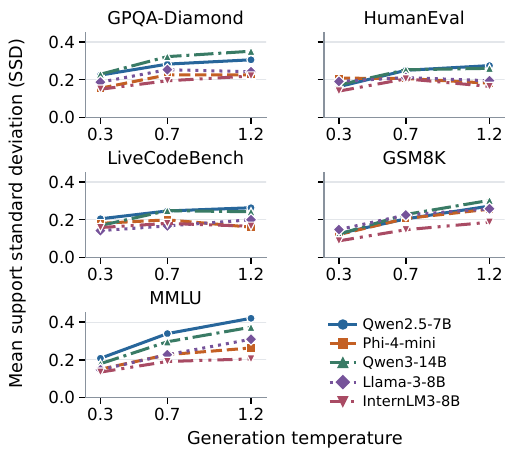}
\caption{\textbf{E2: Support variability across temperatures.} Points show mean question-level SSD (nats per token) for five models, with questions matched across temperatures within each model--benchmark setting.}
\Description{Five panels show mean support standard deviation at generation temperatures 0.3, 0.7, and 1.2 for five models. SSD rises from the lowest to the highest temperature in 23 of 25 settings; Phi-4-mini on HumanEval and LiveCodeBench shows decreases. Several curves decline between the intermediate and highest temperatures.}
\label{fig:pals-e2}
\end{figure}

A \emph{single-round assessment} scores each valid target in one fixed reasoning trace under both the full and deleted contexts. For the set $\mathcal I$ of $m$ valid targets, define average support (AS) and negative support (NS):
\begin{equation}
\label{eq:pals-support-summaries}
\mathrm{AS}=\frac1m\sum_{i\in\mathcal I}s_i,\qquad
\mathrm{NS}=\frac1m\sum_{i\in\mathcal I}(-s_i)_+,
\end{equation}
where $(x)_+=\max(x,0)$. AS measures mean support, while NS averages the magnitudes of negative scores over all $m$ valid targets, assigning zero to nonnegative scores. Both summaries exclude the first step and any separate final-answer block. Traces with no valid targets are omitted. These single-round summaries describe support levels within a fixed trace. Measuring support variability across generated samples requires repeated sampling.

When repeated reasoning samples are available, these summaries also characterize support variability across samples. For $R\geq2$ samples obtained under the same evaluation conditions, let $\mathrm{AS}_r$ and $\mathrm{NS}_r$ denote the support summaries of sample $r$, and let $\overline{\mathrm{AS}}=R^{-1}\sum_r\mathrm{AS}_r$. Define mean negative support and support standard deviation (SSD) as
\begin{equation}
\label{eq:pals-repeat-summaries}
\overline{\mathrm{NS}}=\frac1R\sum_{r=1}^{R}\mathrm{NS}_r,\quad
\mathrm{SSD}=\sqrt{\frac1{R-1}\sum_{r=1}^{R}
(\mathrm{AS}_r-\overline{\mathrm{AS}})^2}.
\end{equation}
For next-step samples from a fixed prefix, each sample contains one target, so $\mathrm{AS}_r=s_r$ and $\mathrm{NS}_r=(-s_r)_+$. Mean negative support summarizes their negative-support burden, while SSD measures variation in support across sampled steps. E2 in Section~\ref{subsec:pals-validation} holds the model and prefix fixed and computes these statistics separately at each generation temperature, with sampling settings unchanged within that temperature.

\subsection{Controlled Validation of PALS}
\label{subsec:pals-validation}
\subsubsection{Constructing reference reasoning DAGs}
Starting from fixed benchmark versions, we obtain reference solutions from official explanations, external model outputs, or explanations of reference programs, applying source-specific answer checks and review. We then construct a DAG by decomposing each solution into self-contained atomic statements, including conditions, background knowledge, derived conclusions, and a separate terminal answer, and at last we annotate direct premise dependencies. Each node retains source evidence and a justification of how its premises support the statement. Avoiding position-dependent references keeps steps interpretable after reordering. Candidate graphs undergo structural checks for acyclicity and connectivity to the answer, together with model review of source fidelity, statement validity, and dependency sufficiency; any revisions are bounded and recorded. Accepted graphs are frozen with provenance and review records. Their non-answer statements, read in the frozen topological order, form the reference CoT; their dependencies guide E1's structural interventions and E2's ancestor prefixes.\looseness=-1

\subsubsection{E1: Response to dependency disruption}
We keep all step texts fixed and vary only their order. From each eligible DAG, we form a \emph{branch-contiguous baseline}: shared premises appear first, and each independent reasoning branch forms a contiguous block in dependency order. This arrangement preserves the DAG dependencies and makes whole-branch exchanges possible. A \emph{legal reorder} swaps two independent branch blocks while retaining their internal order; a \emph{disruptive reorder} swaps two adjacent steps linked by a direct dependency, placing the conclusion before its premise. The tested model scores these fixed texts without generating new steps. We compute AS and NS using the same questions and the intersection of scorable target steps across all three versions. Table~\ref{tab:pals-e1} reports averaged results.

\paragraph{Results and interpretation.} Across all 25 model--benchmark settings, disruptive reordering yields lower AS and higher NS than both the branch-contiguous baseline and legal reordering. For example, Qwen3-14B on GPQA-Diamond shows $\Delta\mathrm{AS}=-0.0640$ and $\Delta\mathrm{NS}=+0.0107$ relative to legal reordering. Legal reordering can raise or lower AS, whereas dependency disruption consistently reduces support with step texts unchanged, supporting PALS's sensitivity to dependency disruption. Expanded paired comparisons against original and branch-contiguous baselines also show AS decreases in all 25 settings for each comparison.

\subsubsection{E2: Support variability across repeatedly sampled next steps}
Higher generation temperatures flatten the token sampling distribution, encouraging more diverse next-step outputs. E2 tests whether this diversity is accompanied by greater variation in PALS support. For each eligible question, a fixed hash-based rule selects a non-answer reference node with parents. Its ancestors, in topological order, form a fixed external prefix ending with a direct parent; the reference target text is withheld. The tested model samples eight next steps from this same prefix at each temperature ($0.3$, $0.7$, and $1.2$), with other sampling settings fixed. Each sampled step is held fixed and assigned a support score using Eq.~\eqref{eq:pals-step-support}. The same probability function is used for scoring at all temperatures, so temperature affects only generation. For each question and temperature, we compute SSD across the eight support scores using Eq.~\eqref{eq:pals-repeat-summaries}. Within each model--benchmark setting, we average these question-level SSD values over the same set of questions with eight valid outputs at all three temperatures, invalid or truncated outputs are excluded. Higher SSD indicates greater variation in support across the sampled next steps (Figure~\ref{fig:pals-e2}).

\paragraph{Results and interpretation.} From $T=0.3$ to $T=1.2$, mean SSD increases in 23 of 25 settings; the exceptions are Phi-4-mini on HumanEval and LiveCodeBench. On MMLU, Qwen3-14B's average support remains nearly unchanged ($0.8492$ versus $0.8486$), while SSD increases from $0.1794$ to $0.3730$, revealing variability concealed by the mean. Negative support captures a different property: on GSM8K, the same model's mean negative support decreases from $0.0067$ to $0.0055$, while SSD rises from $0.1236$ to $0.3030$. Appendix~\ref{app:pals-e2-repeats} reports AS and NS by temperature and repetition. These results support PALS's sensitivity to changes in local support relationships across repeatedly sampled next steps. Several curves decline from $T=0.7$ to $T=1.2$, so endpoint increases do not imply monotonicity. E2 thus validates sensitivity to support variability under fixed prefixes, without assuming that higher temperature always implies less stable support.

\section{CoEvalChain: Collaborative LLM Evaluation}
\label{sec:coevalchain}

\subsection{Framework of CoEvalChain}
The lower half of Figure~\ref{fig:pals-coevalchain-overview} organizes CoEvalChain into four stages. \emph{Specify \& Assign} defines the model, benchmark, and evaluation protocol, selects evaluators based on reputation, participation history, and diversity, and assigns tasks stratified by difficulty with redundant coverage (Appendices~\ref{section:node-selection-sction} and~\ref{section:task-partition-sction}). \emph{Evaluate \& Submit} coordinates execution across participants and binds reports through commit--reveal. \emph{Filter \& Aggregate} applies median/MAD filtering and aggregates retained benchmark reports into one score per model--benchmark pair in each round. \emph{Report \& Audit} reports the mean, standard deviation, and confidence interval from repeated rounds, together with rankings, PALS diagnostics, and audit records. PALS supplies complementary process measurements without changing the benchmark score. Blockchain-backed coordination supports these stages through commitments, verification, incentives, and audit records, without independently guaranteeing correct off-chain execution.\looseness=-1

\subsection{Consensus based on Blockchain}
The requirements in Section~\ref{subsec:collaboration-motivation} motivate a shared protocol for coordinating submissions and verifying reports across participants.

Participation in evaluation tasks requires token staking, imposing a financial barrier to maintaining multiple evaluator identities. This requirement provides an economic deterrent to Sybil participation, while the accuracy of submitted evaluations remains dependent on additional verification and assumptions about adversarial control.

We motivate our consensus logic using Schelling's notion of focal points~\citep{schelling1980strategy}. We model evaluation as a coordination game in which an honestly obtained score is the intended focal point. This is a design assumption; focal-point reasoning alone does not establish a truthful equilibrium.

The selection of a Schelling Point in our scenario is predicated on the expectation that other evaluation results will also converge on a specific score. Given a set of available evaluation result $S$ for all participants, the focal point $s^* \in S$ is achieved when salience is maximized:
\begin{equation}\text{Schelling Point } s^* = \arg\max_{s \in S} P(s), 
\end{equation}
where $P(s)$ represents the probability that evaluation result $s$ is perceived as “unique” or “salient” by the collective.

To limit copying and adaptive reporting based on other evaluators' disclosed results, we employ a \textit{two-phase Commit--Reveal scheme} that binds each evaluator to a score before disclosure.

\begin{itemize}[topsep=2pt,itemsep=0pt,parsep=0pt,partopsep=0pt,leftmargin=1.1em,labelsep=.35em]
    \item \textbf{Commit Phase:} Upon evaluating the model, evaluator $i$ generates a score $s_i$ and a random cryptographic salt $r_i$. Instead of broadcasting the score directly, the evaluator computes a commitment hash $C_i = \text{Hash}(s_i \parallel r_i)$ and broadcasts $C_i$ to the ledger. It ensures the score $s_i$ remains secret, preventing information leakage while locking the evaluator's decision.
    \item \textbf{Reveal Phase:} Once the commitment window closes, the protocol enters into the reveal phase. Evaluators broadcast their original parameters $(s_i, r_i)$, then the smart contract validates that $\text{Hash}(s_i \parallel r_i)$ matches the previously recorded $C_i$. After that, $s_i$ can be accepted in the final candidate set for consensus calculation.

\end{itemize}

\begin{table*}[!ht]
\centering
\fontsize{6.2}{7.6}\selectfont
\setlength{\tabcolsep}{0.8pt}
\renewcommand{\arraystretch}{1.20}
\caption{\textbf{Outcome statistics and process support across models and benchmarks.} Mean, std, and 95\% CI compare decentralized and centralized evaluation. E3 measures AS (nats per token) on shared external references (Ref.) and model-generated traces (Gen.), using each setting's eligible questions. AS cells span both method rows. Detailed outcome data are in Appendix~\ref{app:detailed-tables}.}
\begin{tabular*}{\textwidth}{@{\extracolsep{\fill}}l l *{5}{ccccc}@{}}
\toprule
\multirow{2}{*}{\textbf{Model}} & \multirow{2}{*}{\textbf{Method}} & \multicolumn{5}{c}{\textbf{GSM8K}} & \multicolumn{5}{c}{\textbf{LiveCodeBench}} & \multicolumn{5}{c}{\textbf{HumanEval}} & \multicolumn{5}{c}{\textbf{GPQA-Diamond}} & \multicolumn{5}{c}{\textbf{MMLU}} \\
\cmidrule(lr){3-7} \cmidrule(lr){8-12} \cmidrule(lr){13-17} \cmidrule(lr){18-22} \cmidrule(lr){23-27}
& & mean & std & 95\% CI & \makecell{AS\\Ref.} & \makecell{AS\\Gen.} & mean & std & 95\% CI & \makecell{AS\\Ref.} & \makecell{AS\\Gen.} & mean & std & 95\% CI & \makecell{AS\\Ref.} & \makecell{AS\\Gen.} & mean & std & 95\% CI & \makecell{AS\\Ref.} & \makecell{AS\\Gen.} & mean & std & 95\% CI & \makecell{AS\\Ref.} & \makecell{AS\\Gen.} \\
\midrule
\multirow{2}{*}{\textbf{Qwen-2.5-7B}} & Decentralized & 91.09 & 0.13 & $\pm$0.090 & \multirow{2}{*}{0.425} & \multirow{2}{*}{0.738} & 18.99 & 0.37 & $\pm$0.261 & \multirow{2}{*}{0.270} & \multirow{2}{*}{0.638} & 84.06 & 0.30 & $\pm$0.216 & \multirow{2}{*}{0.327} & \multirow{2}{*}{0.971} & 34.60 & 0.88 & $\pm$0.627 & \multirow{2}{*}{0.271} & \multirow{2}{*}{0.394} & 72.46 & 0.10 & $\pm$0.077 & \multirow{2}{*}{0.453} & \multirow{2}{*}{0.455} \\
 & Centralized & 91.35 & 0.15 & $\pm$0.113 &  &  & 18.45 & 1.57 & $\pm$1.120 &  &  & 83.60 & 1.92 & $\pm$1.371 &  &  & 34.85 & 1.94 & $\pm$1.391 &  &  & 74.48 & 0.16 & $\pm$0.116 &  &  \\
\midrule
\multirow{2}{*}{\textbf{InternLM-3-8B}} & Decentralized & 85.51 & 0.17 & $\pm$0.125 & \multirow{2}{*}{0.277} & \multirow{2}{*}{0.533} & 16.53 & 0.20 & $\pm$0.144 & \multirow{2}{*}{0.190} & \multirow{2}{*}{0.209} & 75.16 & 0.39 & $\pm$0.279 & \multirow{2}{*}{0.225} & \multirow{2}{*}{0.346} & 36.78 & 0.51 & $\pm$0.367 & \multirow{2}{*}{0.215} & \multirow{2}{*}{0.305} & 71.90 & 0.45 & $\pm$0.321 & \multirow{2}{*}{0.267} & \multirow{2}{*}{0.368} \\
 & Centralized & 84.65 & 0.32 & $\pm$0.232 &  &  & 15.56 & 1.31 & $\pm$0.939 &  &  & 71.30 & 1.38 & $\pm$0.999 &  &  & 37.16 & 2.83 & $\pm$2.027 &  &  & 71.87 & 0.41 & $\pm$0.296 &  &  \\
\midrule
\multirow{2}{*}{\textbf{Llama-3-8B}} & Decentralized & 79.68 & 0.18 & $\pm$0.133 & \multirow{2}{*}{0.309} & \multirow{2}{*}{0.771} & 12.75 & 0.20 & $\pm$0.142 & \multirow{2}{*}{0.211} & \multirow{2}{*}{0.538} & 60.89 & 0.49 & $\pm$0.352 & \multirow{2}{*}{0.261} & \multirow{2}{*}{0.705} & 33.31 & 1.27 & $\pm$0.909 & \multirow{2}{*}{0.239} & \multirow{2}{*}{0.416} & 63.44 & 0.14 & $\pm$0.100 & \multirow{2}{*}{0.347} & \multirow{2}{*}{0.433} \\
 & Centralized & 79.59 & 0.71 & $\pm$0.509 &  &  & 12.48 & 0.86 & $\pm$0.617 &  &  & 59.95 & 2.63 & $\pm$1.882 &  &  & 32.69 & 2.60 & $\pm$1.859 &  &  & 63.78 & 0.15 & $\pm$0.111 &  &  \\
\midrule
\multirow{2}{*}{\textbf{Phi-4-Mini}} & Decentralized & 84.48 & 0.66 & $\pm$0.473 & \multirow{2}{*}{0.391} & \multirow{2}{*}{0.597} & 17.10 & 0.28 & $\pm$0.197 & \multirow{2}{*}{0.224} & \multirow{2}{*}{0.533} & 68.91 & 0.95 & $\pm$0.680 & \multirow{2}{*}{0.241} & \multirow{2}{*}{0.581} & 34.28 & 1.36 & $\pm$0.976 & \multirow{2}{*}{0.237} & \multirow{2}{*}{0.355} & 68.50 & 0.32 & $\pm$0.227 & \multirow{2}{*}{0.393} & \multirow{2}{*}{0.394} \\
 & Centralized & 84.59 & 0.40 & $\pm$0.287 &  &  & 19.06 & 1.75 & $\pm$1.249 &  &  & 64.81 & 2.07 & $\pm$1.481 &  &  & 35.59 & 2.83 & $\pm$2.026 &  &  & 71.59 & 0.32 & $\pm$0.227 &  &  \\
\midrule
\multirow{2}{*}{\textbf{Qwen-3-14B}} & Decentralized & 95.88 & 0.14 & $\pm$0.098 & \multirow{2}{*}{0.735} & \multirow{2}{*}{1.084} & 43.45 & 0.48 & $\pm$0.341 & \multirow{2}{*}{0.434} & \multirow{2}{*}{0.553} & 93.54 & 0.44 & $\pm$0.312 & \multirow{2}{*}{0.483} & \multirow{2}{*}{0.891} & 59.38 & 0.92 & $\pm$0.656 & \multirow{2}{*}{0.414} & \multirow{2}{*}{0.417} & 82.63 & 0.17 & $\pm$0.125 & \multirow{2}{*}{0.801} & \multirow{2}{*}{0.714} \\
 & Centralized & 96.14 & 0.25 & $\pm$0.176 &  &  & 54.71 & 3.29 & $\pm$2.355 &  &  & 95.79 & 1.60 & $\pm$1.143 &  &  & 59.76 & 4.25 & $\pm$3.040 &  &  & 84.99 & 0.20 & $\pm$0.145 &  &  \\
\bottomrule
\end{tabular*}
\label{tab:model_performance}
\end{table*}

\begin{figure*}[!t]
\begin{minipage}[t]{\columnwidth}
\vspace{0pt}
\centering
        \includegraphics[width=0.86\linewidth]{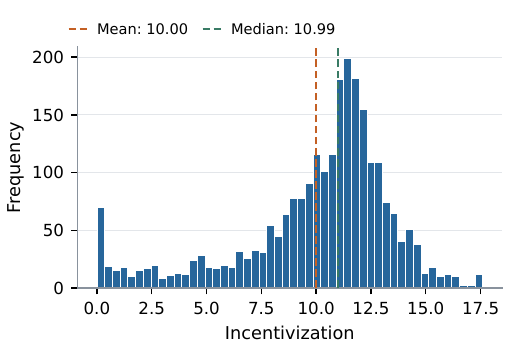}
        \caption{\textbf{Evaluator reward distribution.} Rewards are pooled across benchmark sessions; dashed lines mark the mean and median.}
        \Description{Histogram of evaluator rewards, with dashed lines marking the mean and median.}
        \label{fig:incentivization}
\end{minipage}\hfill
\begin{minipage}[t]{\columnwidth}
\vspace{0pt}
\centering
        \includegraphics[width=0.86\linewidth]{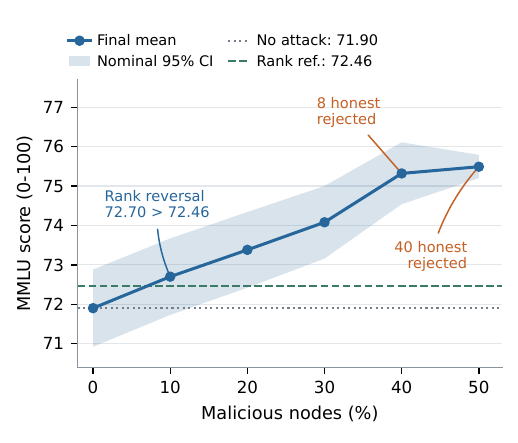}
        \caption{\textbf{Collusion effects on MMLU.} Points show filtered mean scores; shading indicates nominal 95\% CIs.}
        \Description{The mean rises from 71.90 to 72.70 at ten percent malicious nodes, exceeding the reference score of 72.46, and reaches 75.49 at fifty percent. The shaded nominal interval narrows. Eight honest reports are rejected at forty percent malicious nodes and forty at fifty percent.}
        \label{fig:rq5-collusion}
\end{minipage}
\end{figure*}

\subsection{Incentivization Mechanism}
To encourage continued participation, CoEvalChain distributes a fixed reward pool according to each submission's proximity to the batch median, a reference less sensitive to outliers than the mean.

For submitted benchmark scores $\mathcal{S}=\{s_1,\dots,s_n\}$, where $s_i$ is the score reported by evaluator $i$, let $M=\text{median}(s_1,\dots,s_n)$. Evaluator $i$ receives a Gaussian weight that decreases with deviation from $M$:
\begin{equation}
w_{i} = \exp\left(-\frac{(s_{i}-M)^{2}}{2\sigma^{2}}\right),
\end{equation}
where $\sigma=k\cdot\text{MAD}$, $\text{MAD}=\text{median}(|s_i-M|)$, and $k\in[1,1.5]$ controls the decay rate. This scale adapts the reward tolerance to the batch's dispersion.

Each task allocates a fixed pool of $R_{total}=100$ tokens, distributed as
\begin{equation}
R_{i} = R_{total} \cdot \frac{w_i}{\sum_{j=1}^{n} w_j}.
\end{equation}

Here, $R_i$ denotes the number of reward tokens allocated to evaluator $i$. Benchmark scores quantify task performance, while PALS scores quantify stepwise predictive support. Normalization keeps the total reward fixed while favoring submissions close to the batch median.




\section{Experiment}

\subsection{Experimental Setup}
\label{subsec:experimental-setup}
\paragraph{Benchmarks, models, and scale.}
We evaluate mathematical reasoning with GSM8K~\citep{gsm8k}, code generation with HumanEval~\citep{humaneval} and LiveCodeBench~\citep{livecodebench}, scientific reasoning with GPQA-Diamond~\citep{refer31}, and broad knowledge with MMLU~\citep{mmlu}. Five open-weight models---Qwen2.5-7B, Qwen3-14B, Phi-4-mini, InternLM3-8B, and Llama3-8B---cover all five benchmarks. Appendix~\ref{app:additional-model-results} extends outcome evaluation to the larger DeepSeek-V3.2 and proprietary Gemini-2.5-Flash on GSM8K, HumanEval, and GPQA-Diamond, and includes a controlled GPU-hardware ablation. The five-model comparison comprises 5,000 full-benchmark-equivalent evaluations across centralized and decentralized protocols, including all repetitions. Overall, 237 conditions span CoEvalChain outcome evaluation and PALS dependency disruption (E1), temperature-controlled sampling (E2), and single-round process--outcome comparison (E3).\looseness=-1

\paragraph{Repeated outcome evaluation.}
For each model--benchmark pair, we record ten round-level scores per method, separately from the 50-round RIP analysis (Section~\ref{sec:introduction}). Each centralized score averages ten full-benchmark evaluations under the same configuration. Each decentralized score follows CoEvalChain's stratified task assignment and collaborative aggregation across inference configurations and NVIDIA H800, A800, and RTX 5090 GPUs. Both methods evaluate every question exactly ten times per round. Mean, std, and 95\% CI are computed across the ten aggregate scores per method. This comparison matches per-question repetition counts and we also isolate the Task-partition protocol, the results appear in Appendix~\ref{app:task-partition-results}.\looseness=-1

\paragraph{Single-round process assessment.}
E3 measures AS in a single scoring round using two types of reasoning traces. Ref. uses external reference traces shared across models, while Gen. uses each model's step-delimited solution generated once per question with greedy decoding. For each question, we hold the trace fixed and score each valid target step using Eq.~\eqref{eq:pals-step-support}. We then average these step scores to obtain the question's AS, as defined in Eq.~\eqref{eq:pals-support-summaries}. For each trace source and model--benchmark setting, the reported AS is the equally weighted mean of these question-level AS values over eligible questions. These process measurements are separate from repeated outcome evaluation and shared by the decentralized and centralized rows in Table~\ref{tab:model_performance}.\looseness=-1

\subsection{Outcome Reliability: Collaborative versus Centralized Evaluation}
\label{subsec:outcome-results}
Table~\ref{tab:model_performance} shows lower round-level standard deviations under decentralized evaluation in 22 of the 25 model--benchmark settings, with one unchanged and two higher. The reductions span all five models and benchmarks. For Qwen3-14B on GPQA-Diamond, std falls from $4.25$ to $0.92$, a $78.4\%$ reduction, while the reported 95\% CI narrows from $\pm3.040$ to $\pm0.656$. These statistics indicate more concentrated aggregate results and improved precision under the collaborative protocol. Additional results for DeepSeek-V3.2 and Gemini-2.5-Flash in Appendix~\ref{app:additional-model-results} (Table~\ref{tab:api_model_performance}) show lower std and narrower intervals in all six settings.\looseness=-1

\begingroup
\emergencystretch=1em
Across five models and five benchmarks, 23 of 25 model--benchmark configurations exhibit answer-correct traces containing negative-support transitions, with this pattern occurring in up to 21.4\% of answer-correct, scorable traces within a configuration (Appendix~\ref{app:negative-support}). Negative support indicates reduced predictive support from the immediately preceding step, without implying that the target step is logically incorrect. Correct final answers thus leave local support differences unobserved. Likewise, lower dispersion of aggregate benchmark scores does not establish more stable support relationships within individual traces. This motivates complementing outcome reliability with process reliability through single-round PALS measurements on the same model--benchmark pairs. We next examine these support profiles and their relationship to repeated-score dispersion.
\par\endgroup

\subsection{Process Reliability: Single-Round Stepwise Evidence Beyond Outcome Scores}
\label{subsec:process-outcome-results}
E3 reveals different support profiles for shared references and model-generated reasoning (Table~\ref{tab:model_performance}). On HumanEval, Qwen3-14B has higher Ref. AS than Qwen2.5-7B ($0.483$ versus $0.327$), but lower Gen. AS ($0.891$ versus $0.971$). This reversal highlights the roles of both the tested model and the reasoning text being scored. Gen. AS additionally depends on the model's ability to produce a step-delimited trace suitable for assessment. InternLM3-8B yields 7,322 scorable traces out of 15,898 questions ($46.06\%$), illustrating how structured reasoning generation and format adherence affect measurement coverage. Gen. AS characterizes the eligible outputs; Ref. AS provides a complementary assessment on externally prepared reasoning.\looseness=-1

Across the 25 model--benchmark settings, Ref. AS and Gen. AS have Spearman correlations of approximately $-0.357$ and $-0.348$ with decentralized std, respectively. These descriptive associations suggest that stronger stepwise support can accompany lower outcome variability, connecting process measurements with repeated evaluation results. However, individual comparisons can show the opposite pattern. On MMLU, Qwen3-14B has a higher Ref. AS than Qwen2.5-7B (0.801 versus 0.453), but also a higher decentralized std (0.17 versus 0.10). Thus, stronger stepwise support does not always coincide with lower outcome variability. AS summarizes predictive support within a trace, whereas std measures variation across repeated outcome scores. AS is therefore not an estimator of std. Instead, single-round PALS supplies step-level evidence that can be examined alongside repeated-score dispersion, without requiring repeated generation of complete reasoning traces.\looseness=-1

GPQA-Diamond shows comparatively weak support and high variability across models and trace sources. For each model, GPQA-Diamond has the highest standard deviation of collaborative evaluation scores among the five benchmarks. It also ranks among the two lowest benchmarks in both Ref. AS and Gen. AS. The pattern on shared reference traces shows that this observation is not confined to model-generated CoTs. One possible explanation is that demanding scientific inference makes dependence on intermediate premises especially consequential, making the connection between stepwise support and outcome variability more apparent. This task-level co-occurrence motivates further testing of that explanation. It also complements the small score gaps and ranking fragility observed among similarly sized models on GPQA-Diamond in Section~\ref{sec:introduction}. Such evaluations benefit from examining stepwise support, repeated-score dispersion, and model separation together.\looseness=-1

\subsection{Procedural Integrity: Effectiveness and Limits of Blockchain-Based Coordination}
\label{subsec:collaboration-mechanism-results}

\subsubsection{Incentivization Analysis}



We simulate reward allocation using the proposed incentivization mechanism. Figure~\ref{fig:incentivization} shows that most rewards lie between $7.5$ and $14.0$ tokens, with a peak around $11.0$--$11.5$ and approximately $200$ observations in the peak bin. Approximately $70$ observations, accounting for about $3\%$ of the total, receive rewards near zero, consistent with the mechanism's strong downweighting of submissions that deviate from their batch median.

These results characterize how the mechanism rewards agreement with the batch reference. Its effectiveness in encouraging accurate reporting depends on the integrity of that reference: coordinated submissions may agree with one another while remaining systematically biased. We therefore examine how coordinated false reporting affects filtering, aggregation, and model ranking.



\subsubsection{Collusion and Ranking Manipulation}
\label{sec:rq5-robustness}

Coordinated false reporting may arise from collusion among distinct evaluators or from a single adversary operating multiple evaluator identities, as in a Sybil attack~\citep{douceur2002sybil}. Both can give correlated reports disproportionate influence over collaborative evaluation. We assess this threat at the report level, conditioning on the fraction of submissions under adversarial control.

Using 100 existing InternLM-3-8B MMLU evaluation scores, we replace $10\%$--$50\%$ of the reports with a common inflated score, then apply robust filtering and aggregate the retained reports. We examine whether these coordinated submissions shift the estimated performance, reverse its ordering relative to a fixed Qwen2.5-7B reference, and distort apparent estimation precision.

In this simulation (Figure~\ref{fig:rq5-collusion}), controlling $10\%$ of reports raises the aggregate from $71.90$ to $72.70$, reversing its ordering against the $72.46$ reference. At $50\%$ corruption, filtering retains 50 malicious and ten honest reports, raising the malicious share to $83.3\%$. The nominal confidence intervals narrow even as the score moves farther from the no-attack baseline, showing that narrower intervals do not imply more accurate evaluation.\looseness=-1

These results clarify the trust assumptions underlying procedural integrity. Staking imposes a financial barrier to maintaining multiple evaluator identities, while Commit--Reveal constrains copying and adaptive reporting after disclosure. Prearranged false reports can nevertheless satisfy the submission protocol. Reliable aggregation therefore requires additional safeguards against concentrated adversarial control and verification of submitted results beyond their agreement with other reports.

\section{Related Work}

\subsection{Blockchain and Large Language Models}

Machine learning systems often centralize training and data management. Blockchain integration offers alternative mechanisms for coordination and accountability \citep{refer08,refer09}. Blockchain applications span digital assets, transaction analysis, and accountable data sharing \citep{refer10,refer11,refer12}. Its integration with LLMs includes decentralized service provision in LLM-Net \citep{refer07}, collaborative AI training in AIArena \citep{refer06}, and blockchain-inspired multi-agent collaboration for root cause analysis in mABC \citep{refer13}. In the other direction, LLMs support blockchain applications through DeFi transaction intent mining \citep{refer14} and the detection of bad practices in smart contracts \citep{refer15}.\looseness=-1

These directions illustrate the opportunities for bidirectional integration between blockchain and LLMs \citep{refer16}. Our focus is reliable LLM evaluation: CoEvalChain uses decentralized coordination to make repeated evaluation and score aggregation auditable. This procedural role complements statistical measurement; blockchain records alone do not establish the correctness of submitted scores.

\begingroup
\subsection{Evaluation of Large Language Models}

Benchmarks and evaluation methods assess LLM quality, safety, and reliability \citep{refer17,refer18}. A central requirement is to define what each evaluation measures. TruthfulQA assesses models' tendency to reproduce human falsehoods \citep{refer04}, while BeaverTails, SG-Bench, and SafeWorld address safety alignment and generalization \citep{refer19,refer20,refer21}. RBench evaluates graduate-level, multidisciplinary reasoning in English and Chinese \citep{refer18}, and BBQ examines social biases in question answering \citep{refer22}.\looseness=-1

Statistical methods aid interpretation: GEM evaluates generated responses using mutual information without gold-standard references \citep{refer23}; other work studies distributional alignment and benchmark quality \citep{refer25,refer29}. Miller estimates uncertainty in benchmark scores and paired model differences \citep{refer30}. These confidence intervals characterize estimation uncertainty, whereas repeated-run standard deviations describe execution-to-execution score variation.\looseness=-1

Variability measures address distinct sources of inconsistency. The Bias--Volatility Framework separates discrimination risk into bias and volatility risks using the mean and variation of a model's stereotype distribution \citep{refer24}. FormatSpread estimates performance ranges across plausible, meaning-preserving prompt formats \citep{sclar2024formatspread}. Semantic entropy groups semantically equivalent generations to measure uncertainty over meanings \citep{kuhn2023semanticuncertainty}, while SelfCheckGPT derives hallucination-detection scores from consistency with additional sampled responses \citep{manakul2023selfcheckgpt}. These measures differ from score variation under a fixed evaluation protocol.\looseness=-1

Human evaluation provides a complementary perspective on output quality and preference alignment. Prior work compares the stochastic properties of human- and machine-generated text \citep{refer26}, while Chatbot Arena and maximum discrepancy competition support direct human comparisons of model outputs \citep{refer27,refer17}. Our work links repeated model comparisons, stepwise support diagnostics, and auditable aggregation.\looseness=-1

\section{Conclusion}
This work connects outcome reliability, process reliability, and procedural integrity through PALS and CoEvalChain.
Our experiments support PALS as a diagnostic of stepwise predictive support and show that collaborative aggregation generally improves the repeatability of reported benchmark estimates.
The observed process--outcome associations further highlight the value of combining PALS diagnostics with repeated outcome measurements within an auditable evaluation workflow.

Several limitations qualify these findings.
PALS measures predictive support between adjacent steps, not logical correctness.
Blockchain-backed coordination supports traceable submissions and auditable reporting, while trustworthy aggregation additionally depends on off-chain execution and participant behavior.
Our collusion analysis clarifies these application-level trust requirements and motivates additional verification of submitted results.
In particular, we will use PALS to examine how stepwise predictive support varies with task difficulty and model reasoning capability.
\par\endgroup

\endgroup

\endgroup

\begingroup
\renewcommand{\bibfont}{\fontsize{6.5}{7.2}\selectfont}
\bibliography{references}
\bibliographystyle{ACM-Reference-Format-initials}
\endgroup

\appendix
\begingroup
\raggedbottom
\linespread{0.93}\selectfont
\setlength{\textfloatsep}{5pt plus 1pt minus 1pt}
\setlength{\floatsep}{5pt plus 1pt minus 1pt}
\setlength{\intextsep}{3pt plus 1pt minus 1pt}
\captionsetup{skip=2pt,belowskip=0pt}

\makeatletter
\patchcmd{\section}
  {-.75\baselineskip \@plus -2\p@ \@minus -.2\p@}
  {-3pt \@plus -.5pt \@minus -.5pt}{}{}
\patchcmd{\section}{.25\baselineskip}{1pt}{}{}
\patchcmd{\subsection}
  {-.75\baselineskip \@plus -2\p@ \@minus -.2\p@}
  {-2pt \@plus -.5pt \@minus -.5pt}{}{}
\patchcmd{\subsection}{.25\baselineskip}{1pt}{}{}
\makeatother

\section{Participant Selection}
\label{section:node-selection-sction}

Participant selection combines quality scoring with diversity-aware Maximal Marginal Relevance (MMR). For candidate $u_i$ with reputation $r_i$ and $t_i$ previous task participations, define its quality score as
\begin{equation}
    q_{i} = \frac{r_{i}}{1 + \gamma \cdot t_{i}},
\end{equation}
Here, $\gamma \ge 0$ controls the penalty for frequent participation, discouraging task concentration among a few active nodes.

Node similarity uses explicit attributes (skill keywords, company affiliation, job position, and educational background) and implicit features (latent themes and areas of expertise). Given the selected set $S$, MMR greedily adds the unselected candidate maximizing
\begin{equation}
    \text{Score} = \lambda \cdot q_{i} - (1 - \lambda) \cdot \max_{u_j \in S} \text{Sim}(u_i, u_j),
\end{equation}
where $\lambda \in [0,1]$ balances quality against similarity to the most similar selected node. Selection continues until $k$ participants have been selected.\looseness=-1

\section{Task Partitioning}
\label{section:task-partition-sction}

For the $N$ evaluators in $\mathcal{V}$ selected by computational capability and historical reliability (Appendix~\ref{section:node-selection-sction}), we partition benchmark $\mathcal{D}$ into $N$ subsets $\{\mathcal{S}_1,\dots,\mathcal{S}_N\}$ using a \textit{Stratified Redundant Partitioning Algorithm}. Workloads adapt dynamically under two constraints:\looseness=-1

\begin{enumerate}
    \item \textbf{Redundant coverage:} Each sample $d\in\mathcal{D}$ appears in exactly $\rho$ distinct subsets assigned to independent evaluators, where $\rho\geq1$. This supports Schelling Point consensus and outlier evaluator detection.
    \item \textbf{Stratified difficulty:} Divide $\mathcal{D}$ into $K$ difficulty strata $\{L_1,\dots,L_K\}$. Each subset $\mathcal{S}_i$ matches $\mathcal{D}$'s difficulty distribution, providing representative task mixtures and supporting fair Proof-of-Work (PoW) contributions.
\end{enumerate}

\section{Negative Support in Answer-Correct Traces}
\label{app:negative-support}

Across five models and five benchmarks, Table~\ref{tab:negative-support} reports the percentage of answer-correct, scorable E3 traces containing at least one negative-support transition ($s_i<0$, equivalently $\mathrm{NS}>0$). Traces are generated once per question with greedy decoding; unscorable traces and those without a determinate correct outcome are excluded. Under this frozen protocol, the pattern appears in 23 of 25 configurations, reaching 21.43\% for Qwen2.5-7B on HumanEval. Correct final answers can therefore coexist with reduced predictive support between adjacent reasoning steps. PALS complements outcome-level accuracy by making these local support differences observable, without equating negative support with a logical error.\looseness=-1

\begin{table}[!ht]
\centering
\caption{\textbf{Answer-correct, scorable E3 traces containing at least one negative-support transition (\%).}}
\label{tab:negative-support}
\begingroup
\footnotesize
\setlength{\tabcolsep}{1.5pt}
\renewcommand{\arraystretch}{1}
\begin{tabular*}{\columnwidth}{@{\extracolsep{\fill}}l*{5}{c}@{}}
\toprule
\textbf{Model} & \textbf{GSM8K} & \textbf{HumanEval} & \makecell{\textbf{LiveCode}\\\textbf{Bench}} & \makecell{\textbf{GPQA-}\\\textbf{Diamond}} & \textbf{MMLU} \\
\midrule
Qwen2.5-7B & 11.74\% & 21.43\% & 12.90\% & 1.56\% & 12.36\% \\
\addlinespace[1pt]
Qwen3-14B & 11.78\% & 5.71\% & 2.00\% & 5.43\% & 18.54\% \\
\addlinespace[1pt]
Phi-4-mini & 2.78\% & 14.13\% & 3.85\% & 6.98\% & 10.10\% \\
\addlinespace[1pt]
Llama-3-8B & 3.24\% & 10.84\% & 9.09\% & 3.13\% & 8.53\% \\
\addlinespace[1pt]
InternLM3-8B & 2.82\% & 3.26\% & 0.00\% & 0.00\% & 1.34\% \\
\bottomrule
\end{tabular*}
\endgroup

\end{table}

\section{Supplementary Evaluation Experiments}
\label{app:additional-model-results}

\subsection{Score Consistency Across Stratified Partitions}
\label{app:task-partition-results}

Table~\ref{tab:task-partition-result} shows closely matched GSM8K scores across partitions, with gaps of $0.17$ and $0.04$ percentage points for Gemini-2.5-Flash and DeepSeek-V3.2, respectively. GPQA-Diamond gaps are larger ($1.84$ and $3.21$ points), while DeepSeek-V3.2 differs by $5.51$ points on HumanEval. Partition consistency therefore remains benchmark dependent despite stratification.\looseness=-1

\begin{table}[!ht]\small
\centering
\setlength{\tabcolsep}{2pt}
\renewcommand{\arraystretch}{1}
\caption{\textbf{Benchmark scores (\%) across stratified partitions.}}
\begin{tabular*}{\columnwidth}{@{\extracolsep{\fill}}l cc cc@{}}
\toprule
\multirow{2}{*}{\textbf{Benchmark}} & \multicolumn{2}{c}{\textbf{Gemini-2.5-Flash}} & \multicolumn{2}{c}{\textbf{Deepseek-V3.2}} \\
\cmidrule(lr){2-3} \cmidrule(lr){4-5} 
& partition-1 & partition-2 & partition-1 & partition-2 \\
\midrule
\textbf{GSM8k} & 94.69 & 94.52 & 95.53 & 95.57 \\
\midrule
\textbf{Humaneval} & 98.29 & 97.04 & 95.95 & 90.44 \\
\midrule
\textbf{GPQA-Diamond} & 76.73 & 78.57 & 77.05 & 73.84 \\
\bottomrule
\end{tabular*}
\label{tab:task-partition-result}
\end{table}

\subsection{Collaborative versus Centralized Evaluation on Additional Models}
\label{app:additional-model-comparison}

Table~\ref{tab:api_model_performance} shows lower std and narrower confidence intervals under collaborative evaluation for both Gemini-2.5-Flash and DeepSeek-V3.2 across all three benchmarks, extending the repeatability findings to these models.\looseness=-1

\begin{table}[!ht]\fontsize{6.2}{7.4}\selectfont
\centering
\setlength{\tabcolsep}{1.3pt}
\renewcommand{\arraystretch}{1.1}
\caption{\textbf{Additional-model outcome statistics.} Mean, std, and 95\% CI across evaluation rounds.}
\begin{tabular*}{\columnwidth}{@{\extracolsep{\fill}}l l ccc ccc ccc@{}}
\toprule
\multirow{2}{*}{\textbf{Model}} & \multirow{2}{*}{\textbf{\makecell[c]{Evaluation \\ Method}}} & \multicolumn{3}{c}{\textbf{GSM8K}} & \multicolumn{3}{c}{\textbf{Humaneval}} & \multicolumn{3}{c}{\textbf{GPQA-Diamond}} \\
\cmidrule(lr){3-5} \cmidrule(lr){6-8} \cmidrule(lr){9-11} 
& & mean & std & 95\% CI & mean & std & 95\% CI & mean & std & 95\% CI \\
\midrule
\multirow{2}{*}{\textbf{Gemini-2.5-Flash}} & Decentralized & 94.61 & 0.31 & $\pm$0.223 & 97.66 & 0.43 & $\pm$0.309 & 77.65 & 0.69 & $\pm$0.494 \\
& Centralized & 94.60 & 0.56 & $\pm$0.403 & 97.86 & 0.77 & $\pm$0.554 & 75.96 & 2.53 & $\pm$1.809 \\
\midrule
\multirow{2}{*}{\textbf{DeepSeek-V3.2}} & Decentralized & 95.56 & 0.29 & $\pm$0.207 & 93.19 & 0.54 & $\pm$0.385  & 75.44 & 0.88 & $\pm$0.630 \\
& Centralized & 95.34 & 0.53 & $\pm$0.379 & 92.86 & 1.22 & $\pm$0.874 & 74.97 & 1.86 & $\pm$1.331 \\
\bottomrule
\end{tabular*}
\label{tab:api_model_performance}
\end{table}

\subsection{Sensitivity to GPU Configuration}
\label{app:hardware-sensitivity}

Table~\ref{tab:hardware-ablation-result} compares Qwen3-14B round-level averages under H800-only and heterogeneous GPU configurations, with all other evaluation settings fixed. H800-only yields lower std than heterogeneous GPU configurations on all three benchmarks. These differences motivate explicit hardware coverage through CoEvalChain, since repeatability under fixed hardware does not establish robustness across configurations.\looseness=-1

\begin{table}[!htb]\footnotesize
\centering
\captionsetup{skip=2pt}
\setlength{\tabcolsep}{2pt}
\renewcommand{\arraystretch}{1}
\caption{\textbf{Hardware sensitivity of Qwen3-14B aggregate scores.}}
\label{tab:hardware-ablation-result}
\begin{tabular*}{\columnwidth}{@{\extracolsep{\fill}}lccc ccc@{}}
\toprule
\multirow{2}{*}{Benchmark} & \multicolumn{3}{c}{H800 only} & \multicolumn{3}{c}{Various GPUs} \\
\cmidrule(lr){2-4}\cmidrule(lr){5-7}
& mean & std & 95\% CI & mean & std & 95\% CI \\
\midrule
GSM8K & 96.27 & 0.64 & $\pm$0.124 & 96.01 & 0.88 & $\pm$0.173 \\
HumanEval & 95.04 & 2.06 & $\pm$0.405 & 94.36 & 2.75 & $\pm$0.540 \\
GPQA-Diamond & 61.73 & 5.43 & $\pm$1.065 & 60.05 & 6.81 & $\pm$1.334 \\
\bottomrule
\end{tabular*}
\end{table}

\clearpage
\endgroup
\begin{table*}[!t]
\centering
\refstepcounter{section}
\label{app:pals-e2-repeats}
\captionsetup{skip=6pt}
\caption{\textbf{Appendix \thesection: E2 AS and NS by Repetition.} Entries are AS / NS (nats per token), averaged across the same complete question cohort at each temperature and repeat index. Repeat 1--8 correspond to raw indices 0--7.}
\label{tab:pals-e2-repeats}
\fontsize{6.8}{7.0}\selectfont
\setlength{\tabcolsep}{2.1pt}
\renewcommand{\arraystretch}{1}
\begin{tabular*}{\textwidth}{@{\extracolsep{\fill}}lc*{8}{c}@{}}
\toprule
Model & $T$ & Repeat 1 & Repeat 2 & Repeat 3 & Repeat 4 & Repeat 5 & Repeat 6 & Repeat 7 & Repeat 8 \\
\midrule
\multicolumn{10}{@{}l}{\textbf{GPQA-Diamond}} \\
\multirow{3}{*}{Qwen2.5-7B} & 0.3 & 0.4825 / 0.0069 & 0.4436 / 0.0073 & 0.4263 / 0.0088 & 0.5008 / 0.0094 & 0.4714 / 0.0147 & 0.4278 / 0.0079 & 0.4575 / 0.0073 & 0.4931 / 0.0084 \\
 & 0.7 & 0.5243 / 0.0076 & 0.4059 / 0.0094 & 0.4531 / 0.0042 & 0.4230 / 0.0085 & 0.4478 / 0.0037 & 0.4951 / 0.0060 & 0.4910 / 0.0088 & 0.4622 / 0.0079 \\
 & 1.2 & 0.4180 / 0.0073 & 0.4166 / 0.0076 & 0.4333 / 0.0048 & 0.4450 / 0.0035 & 0.4433 / 0.0070 & 0.4698 / 0.0072 & 0.4762 / 0.0072 & 0.4422 / 0.0022 \\
\multirow{3}{*}{Phi-4-mini} & 0.3 & 0.3624 / 0.0035 & 0.3216 / 0.0003 & 0.3189 / 0.0016 & 0.4040 / 0.0013 & 0.3717 / 0.0013 & 0.3489 / 0.0016 & 0.3795 / 0.0016 & 0.3339 / 0.0014 \\
 & 0.7 & 0.3455 / 0.0042 & 0.3409 / 0.0028 & 0.3317 / 0.0073 & 0.3538 / 0.0033 & 0.3837 / 0.0001 & 0.4145 / 0.0036 & 0.3157 / 0.0017 & 0.3112 / 0.0022 \\
 & 1.2 & 0.3466 / 0.0032 & 0.2955 / 0.0018 & 0.2945 / 0.0032 & 0.2892 / 0.0068 & 0.2821 / 0.0038 & 0.2738 / 0.0047 & 0.2569 / 0.0049 & 0.2567 / 0.0006 \\
\multirow{3}{*}{Qwen3-14B} & 0.3 & 0.5610 / 0.0041 & 0.5814 / 0.0021 & 0.5788 / 0.0044 & 0.5849 / 0.0076 & 0.5400 / 0.0043 & 0.6239 / 0.0031 & 0.5682 / 0.0046 & 0.5422 / 0.0036 \\
 & 0.7 & 0.5939 / 0.0016 & 0.6091 / 0.0022 & 0.6143 / 0.0022 & 0.6059 / 0.0018 & 0.5433 / 0.0009 & 0.6408 / 0.0024 & 0.5795 / 0.0038 & 0.5435 / 0.0012 \\
 & 1.2 & 0.6391 / 0.0015 & 0.5373 / 0.0056 & 0.6384 / 0.0041 & 0.5424 / 0.0056 & 0.5591 / 0.0044 & 0.4400 / 0.0034 & 0.6043 / 0.0046 & 0.6116 / 0.0016 \\
\multirow{3}{*}{Llama-3-8B} & 0.3 & 0.3790 / 0.0033 & 0.3694 / 0.0059 & 0.4231 / 0.0041 & 0.4144 / 0.0045 & 0.3705 / 0.0034 & 0.3334 / 0.0025 & 0.3808 / 0.0026 & 0.3630 / 0.0033 \\
 & 0.7 & 0.4540 / 0.0001 & 0.4241 / 0.0025 & 0.3263 / 0.0036 & 0.4531 / 0.0025 & 0.3374 / 0.0056 & 0.3551 / 0.0014 & 0.3548 / 0.0039 & 0.4269 / 0.0015 \\
 & 1.2 & 0.2692 / 0.0153 & 0.2965 / 0.0110 & 0.3617 / 0.0015 & 0.2949 / 0.0043 & 0.2772 / 0.0013 & 0.3661 / 0.0042 & 0.3011 / 0.0032 & 0.3104 / 0.0015 \\
\multirow{3}{*}{InternLM3-8B} & 0.3 & 0.3318 / 0.0055 & 0.3613 / 0.0062 & 0.3886 / 0.0070 & 0.3700 / 0.0034 & 0.3733 / 0.0076 & 0.3748 / 0.0047 & 0.3830 / 0.0048 & 0.3696 / 0.0043 \\
 & 0.7 & 0.3288 / 0.0031 & 0.3481 / 0.0027 & 0.2967 / 0.0036 & 0.3215 / 0.0037 & 0.3425 / 0.0042 & 0.3179 / 0.0073 & 0.3511 / 0.0103 & 0.3396 / 0.0052 \\
 & 1.2 & 0.2913 / 0.0028 & 0.2531 / 0.0019 & 0.2712 / 0.0047 & 0.2433 / 0.0064 & 0.2787 / 0.0035 & 0.2616 / 0.0057 & 0.1977 / 0.0061 & 0.2855 / 0.0106 \\
\midrule
\multicolumn{10}{@{}l}{\textbf{HumanEval}} \\
\multirow{3}{*}{Qwen2.5-7B} & 0.3 & 0.4337 / 0.0037 & 0.4999 / 0.0031 & 0.4242 / 0.0006 & 0.4743 / 0.0000 & 0.5135 / 0.0005 & 0.5289 / 0.0014 & 0.4467 / 0.0012 & 0.4415 / 0.0013 \\
 & 0.7 & 0.4892 / 0.0006 & 0.4793 / 0.0021 & 0.4605 / 0.0031 & 0.4657 / 0.0022 & 0.5175 / 0.0000 & 0.4332 / 0.0036 & 0.5156 / 0.0002 & 0.5133 / 0.0033 \\
 & 1.2 & 0.4692 / 0.0018 & 0.5217 / 0.0010 & 0.4579 / 0.0007 & 0.4929 / 0.0023 & 0.4502 / 0.0000 & 0.4390 / 0.0023 & 0.5556 / 0.0018 & 0.5382 / 0.0007 \\
\multirow{3}{*}{Phi-4-mini} & 0.3 & 0.3393 / 0.0010 & 0.2993 / 0.0000 & 0.2638 / 0.0079 & 0.3833 / 0.0052 & 0.3522 / 0.0033 & 0.3893 / 0.0035 & 0.4032 / 0.0012 & 0.4378 / 0.0008 \\
 & 0.7 & 0.2751 / 0.0017 & 0.3668 / 0.0008 & 0.3165 / 0.0076 & 0.2788 / 0.0132 & 0.3600 / 0.0000 & 0.3765 / 0.0012 & 0.2517 / 0.0007 & 0.3277 / 0.0000 \\
 & 1.2 & 0.1687 / 0.0025 & 0.1719 / 0.0133 & 0.2474 / 0.0095 & 0.1881 / 0.0016 & 0.2254 / 0.0000 & 0.2747 / 0.0000 & 0.2632 / 0.0061 & 0.1573 / 0.0060 \\
\multirow{3}{*}{Qwen3-14B} & 0.3 & 0.5273 / 0.0014 & 0.5308 / 0.0009 & 0.5163 / 0.0006 & 0.5019 / 0.0021 & 0.5268 / 0.0011 & 0.5394 / 0.0010 & 0.5669 / 0.0017 & 0.5446 / 0.0022 \\
 & 0.7 & 0.4623 / 0.0001 & 0.4927 / 0.0029 & 0.4959 / 0.0088 & 0.5369 / 0.0022 & 0.4521 / 0.0013 & 0.5307 / 0.0018 & 0.5332 / 0.0015 & 0.5238 / 0.0007 \\
 & 1.2 & 0.4810 / 0.0023 & 0.5074 / 0.0010 & 0.5071 / 0.0113 & 0.4928 / 0.0048 & 0.4826 / 0.0016 & 0.4765 / 0.0088 & 0.4146 / 0.0011 & 0.5030 / 0.0043 \\
\multirow{3}{*}{Llama-3-8B} & 0.3 & 0.2574 / 0.0095 & 0.2536 / 0.0018 & 0.3411 / 0.0007 & 0.2805 / 0.0053 & 0.2448 / 0.0023 & 0.3327 / 0.0018 & 0.3188 / 0.0072 & 0.2335 / 0.0017 \\
 & 0.7 & 0.2826 / 0.0268 & 0.2597 / 0.0131 & 0.2947 / 0.0028 & 0.2321 / 0.0046 & 0.2483 / 0.0028 & 0.2616 / 0.0033 & 0.1872 / 0.0017 & 0.1907 / 0.0047 \\
 & 1.2 & 0.0795 / 0.0391 & 0.2030 / 0.0106 & 0.1270 / 0.0108 & 0.2366 / 0.0133 & 0.1553 / 0.0184 & 0.2098 / 0.0075 & 0.0757 / 0.0057 & 0.1472 / 0.0018 \\
\multirow{3}{*}{InternLM3-8B} & 0.3 & 0.3883 / 0.0062 & 0.4238 / 0.0047 & 0.4052 / 0.0066 & 0.3876 / 0.0023 & 0.3933 / 0.0077 & 0.3920 / 0.0053 & 0.4138 / 0.0034 & 0.3328 / 0.0067 \\
 & 0.7 & 0.4083 / 0.0030 & 0.2645 / 0.0031 & 0.3180 / 0.0014 & 0.3394 / 0.0033 & 0.4259 / 0.0019 & 0.3730 / 0.0029 & 0.3477 / 0.0066 & 0.3912 / 0.0017 \\
 & 1.2 & 0.2651 / 0.0013 & 0.2257 / 0.0081 & 0.2272 / 0.0003 & 0.2549 / 0.0000 & 0.2104 / 0.0072 & 0.1964 / 0.0079 & 0.1908 / 0.0024 & 0.2688 / 0.0008 \\
\midrule
\multicolumn{10}{@{}l}{\textbf{LiveCodeBench v6}} \\
\multirow{3}{*}{Qwen2.5-7B} & 0.3 & 0.4214 / 0.0101 & 0.4029 / 0.0076 & 0.3903 / 0.0103 & 0.3657 / 0.0111 & 0.4447 / 0.0085 & 0.3935 / 0.0081 & 0.4292 / 0.0070 & 0.3923 / 0.0096 \\
 & 0.7 & 0.3487 / 0.0130 & 0.4199 / 0.0071 & 0.3853 / 0.0092 & 0.3703 / 0.0070 & 0.4116 / 0.0043 & 0.3922 / 0.0041 & 0.4395 / 0.0086 & 0.4455 / 0.0056 \\
 & 1.2 & 0.3774 / 0.0016 & 0.4108 / 0.0046 & 0.3347 / 0.0052 & 0.3685 / 0.0014 & 0.4446 / 0.0021 & 0.3346 / 0.0136 & 0.4487 / 0.0051 & 0.3661 / 0.0112 \\
\multirow{3}{*}{Phi-4-mini} & 0.3 & 0.2871 / 0.0097 & 0.2997 / 0.0133 & 0.3026 / 0.0140 & 0.3750 / 0.0107 & 0.2987 / 0.0126 & 0.2821 / 0.0153 & 0.3096 / 0.0176 & 0.2905 / 0.0084 \\
 & 0.7 & 0.2170 / 0.0090 & 0.2926 / 0.0125 & 0.2430 / 0.0063 & 0.2572 / 0.0071 & 0.3199 / 0.0080 & 0.2467 / 0.0046 & 0.2908 / 0.0085 & 0.2278 / 0.0112 \\
 & 1.2 & 0.1584 / 0.0018 & 0.1581 / 0.0124 & 0.1745 / 0.0050 & 0.1959 / 0.0014 & 0.1920 / 0.0088 & 0.2289 / 0.0032 & 0.1775 / 0.0006 & 0.1461 / 0.0111 \\
\multirow{3}{*}{Qwen3-14B} & 0.3 & 0.3724 / 0.0046 & 0.4082 / 0.0047 & 0.3786 / 0.0079 & 0.3967 / 0.0074 & 0.3571 / 0.0081 & 0.4035 / 0.0072 & 0.4212 / 0.0081 & 0.3970 / 0.0100 \\
 & 0.7 & 0.4187 / 0.0057 & 0.4216 / 0.0016 & 0.4055 / 0.0063 & 0.4344 / 0.0077 & 0.3660 / 0.0077 & 0.3925 / 0.0068 & 0.4415 / 0.0061 & 0.4486 / 0.0054 \\
 & 1.2 & 0.4323 / 0.0037 & 0.3815 / 0.0110 & 0.4183 / 0.0053 & 0.3868 / 0.0030 & 0.4451 / 0.0047 & 0.3924 / 0.0094 & 0.4143 / 0.0051 & 0.4447 / 0.0050 \\
\multirow{3}{*}{Llama-3-8B} & 0.3 & 0.2364 / 0.0045 & 0.2673 / 0.0057 & 0.1547 / 0.0109 & 0.2464 / 0.0052 & 0.2743 / 0.0050 & 0.2282 / 0.0106 & 0.2084 / 0.0140 & 0.1784 / 0.0075 \\
 & 0.7 & 0.2011 / 0.0056 & 0.2483 / 0.0054 & 0.2265 / 0.0042 & 0.2136 / 0.0064 & 0.1801 / 0.0036 & 0.2706 / 0.0010 & 0.1859 / 0.0051 & 0.2374 / 0.0035 \\
 & 1.2 & 0.1262 / 0.0029 & 0.1553 / 0.0041 & 0.1649 / 0.0402 & 0.1663 / 0.0026 & 0.2108 / 0.0004 & 0.2509 / 0.0031 & 0.1531 / 0.0054 & 0.2449 / 0.0023 \\
\multirow{3}{*}{InternLM3-8B} & 0.3 & 0.2898 / 0.0058 & 0.2306 / 0.0058 & 0.2629 / 0.0075 & 0.2356 / 0.0021 & 0.2867 / 0.0051 & 0.2566 / 0.0040 & 0.2812 / 0.0072 & 0.2787 / 0.0009 \\
 & 0.7 & 0.2536 / 0.0030 & 0.1963 / 0.0049 & 0.2865 / 0.0020 & 0.2798 / 0.0003 & 0.2797 / 0.0016 & 0.2623 / 0.0006 & 0.2224 / 0.0028 & 0.2985 / 0.0039 \\
 & 1.2 & 0.2125 / 0.0010 & 0.1665 / 0.0011 & 0.2063 / 0.0031 & 0.2855 / 0.0008 & 0.2042 / 0.0010 & 0.2315 / 0.0007 & 0.2355 / 0.0013 & 0.1839 / 0.0094 \\
\midrule
\multicolumn{10}{@{}l}{\textbf{GSM8K}} \\
\multirow{3}{*}{Qwen2.5-7B} & 0.3 & 0.6707 / 0.0078 & 0.6637 / 0.0099 & 0.6750 / 0.0080 & 0.6660 / 0.0082 & 0.6492 / 0.0087 & 0.6660 / 0.0085 & 0.6573 / 0.0109 & 0.6524 / 0.0097 \\
 & 0.7 & 0.6477 / 0.0071 & 0.6821 / 0.0098 & 0.6483 / 0.0092 & 0.6567 / 0.0086 & 0.6717 / 0.0085 & 0.6531 / 0.0081 & 0.6637 / 0.0077 & 0.6657 / 0.0092 \\
 & 1.2 & 0.6612 / 0.0086 & 0.6294 / 0.0096 & 0.6550 / 0.0103 & 0.6331 / 0.0093 & 0.6433 / 0.0093 & 0.6375 / 0.0074 & 0.6706 / 0.0065 & 0.6359 / 0.0114 \\
\multirow{3}{*}{Phi-4-mini} & 0.3 & 0.6116 / 0.0042 & 0.6011 / 0.0040 & 0.6181 / 0.0046 & 0.6220 / 0.0041 & 0.6333 / 0.0033 & 0.6044 / 0.0056 & 0.6406 / 0.0046 & 0.6265 / 0.0040 \\
 & 0.7 & 0.6008 / 0.0047 & 0.6017 / 0.0058 & 0.6054 / 0.0033 & 0.5958 / 0.0058 & 0.5921 / 0.0054 & 0.5980 / 0.0044 & 0.5985 / 0.0038 & 0.6109 / 0.0043 \\
 & 1.2 & 0.5320 / 0.0051 & 0.5373 / 0.0051 & 0.5074 / 0.0066 & 0.5580 / 0.0059 & 0.5239 / 0.0072 & 0.5270 / 0.0028 & 0.5338 / 0.0062 & 0.5497 / 0.0058 \\
\multirow{3}{*}{Qwen3-14B} & 0.3 & 0.9448 / 0.0075 & 0.9411 / 0.0074 & 0.9283 / 0.0055 & 0.9425 / 0.0064 & 0.9582 / 0.0065 & 0.9463 / 0.0068 & 0.9515 / 0.0064 & 0.9227 / 0.0068 \\
 & 0.7 & 0.9344 / 0.0059 & 0.9239 / 0.0070 & 0.9317 / 0.0056 & 0.9473 / 0.0070 & 0.9478 / 0.0055 & 0.9118 / 0.0066 & 0.9299 / 0.0055 & 0.9272 / 0.0074 \\
 & 1.2 & 0.8870 / 0.0054 & 0.9112 / 0.0055 & 0.9109 / 0.0061 & 0.9353 / 0.0047 & 0.8983 / 0.0054 & 0.9180 / 0.0048 & 0.9326 / 0.0064 & 0.9015 / 0.0060 \\
\multirow{3}{*}{Llama-3-8B} & 0.3 & 0.5250 / 0.0066 & 0.5125 / 0.0074 & 0.5297 / 0.0065 & 0.5335 / 0.0045 & 0.5445 / 0.0055 & 0.5044 / 0.0053 & 0.5070 / 0.0073 & 0.5132 / 0.0065 \\
 & 0.7 & 0.4959 / 0.0042 & 0.5048 / 0.0051 & 0.5119 / 0.0063 & 0.5112 / 0.0052 & 0.5187 / 0.0061 & 0.5158 / 0.0057 & 0.5304 / 0.0055 & 0.5320 / 0.0055 \\
 & 1.2 & 0.4595 / 0.0062 & 0.4663 / 0.0046 & 0.4829 / 0.0028 & 0.4645 / 0.0029 & 0.4550 / 0.0130 & 0.4906 / 0.0048 & 0.4648 / 0.0045 & 0.5028 / 0.0044 \\
\multirow{3}{*}{InternLM3-8B} & 0.3 & 0.4134 / 0.0017 & 0.4151 / 0.0029 & 0.4148 / 0.0026 & 0.4141 / 0.0025 & 0.4170 / 0.0027 & 0.4127 / 0.0024 & 0.4152 / 0.0031 & 0.4155 / 0.0029 \\
 & 0.7 & 0.4051 / 0.0030 & 0.4058 / 0.0026 & 0.3941 / 0.0028 & 0.4013 / 0.0027 & 0.3962 / 0.0033 & 0.4080 / 0.0027 & 0.4190 / 0.0037 & 0.4073 / 0.0040 \\
 & 1.2 & 0.3525 / 0.0030 & 0.3745 / 0.0032 & 0.3621 / 0.0037 & 0.3603 / 0.0025 & 0.3760 / 0.0031 & 0.3579 / 0.0043 & 0.3762 / 0.0033 & 0.3625 / 0.0022 \\
\midrule
\multicolumn{10}{@{}l}{\textbf{MMLU}} \\
\multirow{3}{*}{Qwen2.5-7B} & 0.3 & 0.7869 / 0.0105 & 0.7828 / 0.0095 & 0.7793 / 0.0095 & 0.7745 / 0.0099 & 0.7824 / 0.0096 & 0.7924 / 0.0096 & 0.7891 / 0.0091 & 0.7793 / 0.0106 \\
 & 0.7 & 0.7710 / 0.0109 & 0.7797 / 0.0120 & 0.7823 / 0.0098 & 0.7806 / 0.0122 & 0.7750 / 0.0117 & 0.7611 / 0.0119 & 0.7653 / 0.0116 & 0.7758 / 0.0114 \\
 & 1.2 & 0.7606 / 0.0129 & 0.7450 / 0.0136 & 0.7450 / 0.0127 & 0.7523 / 0.0124 & 0.7437 / 0.0131 & 0.7485 / 0.0114 & 0.7608 / 0.0125 & 0.7423 / 0.0122 \\
\multirow{3}{*}{Phi-4-mini} & 0.3 & 0.4683 / 0.0043 & 0.4713 / 0.0054 & 0.4616 / 0.0053 & 0.4710 / 0.0051 & 0.4592 / 0.0058 & 0.4676 / 0.0051 & 0.4677 / 0.0046 & 0.4638 / 0.0051 \\
 & 0.7 & 0.4325 / 0.0072 & 0.4458 / 0.0063 & 0.4521 / 0.0062 & 0.4403 / 0.0061 & 0.4423 / 0.0070 & 0.4482 / 0.0069 & 0.4488 / 0.0069 & 0.4437 / 0.0056 \\
 & 1.2 & 0.3879 / 0.0086 & 0.3864 / 0.0086 & 0.3804 / 0.0076 & 0.3882 / 0.0076 & 0.3821 / 0.0062 & 0.3818 / 0.0070 & 0.3896 / 0.0077 & 0.3875 / 0.0077 \\
\multirow{3}{*}{Qwen3-14B} & 0.3 & 0.8505 / 0.0053 & 0.8474 / 0.0052 & 0.8497 / 0.0052 & 0.8554 / 0.0053 & 0.8580 / 0.0050 & 0.8411 / 0.0048 & 0.8426 / 0.0050 & 0.8490 / 0.0058 \\
 & 0.7 & 0.8472 / 0.0048 & 0.8544 / 0.0051 & 0.8412 / 0.0061 & 0.8588 / 0.0047 & 0.8625 / 0.0054 & 0.8446 / 0.0051 & 0.8363 / 0.0058 & 0.8622 / 0.0055 \\
 & 1.2 & 0.8480 / 0.0053 & 0.8490 / 0.0062 & 0.8457 / 0.0063 & 0.8444 / 0.0054 & 0.8528 / 0.0058 & 0.8542 / 0.0055 & 0.8537 / 0.0055 & 0.8407 / 0.0052 \\
\multirow{3}{*}{Llama-3-8B} & 0.3 & 0.4394 / 0.0028 & 0.4389 / 0.0023 & 0.4274 / 0.0030 & 0.4237 / 0.0027 & 0.4376 / 0.0025 & 0.4272 / 0.0028 & 0.4396 / 0.0029 & 0.4361 / 0.0025 \\
 & 0.7 & 0.4282 / 0.0036 & 0.4291 / 0.0032 & 0.4237 / 0.0034 & 0.4299 / 0.0029 & 0.4273 / 0.0028 & 0.4309 / 0.0034 & 0.4377 / 0.0028 & 0.4334 / 0.0033 \\
 & 1.2 & 0.4239 / 0.0057 & 0.4319 / 0.0047 & 0.4180 / 0.0055 & 0.4173 / 0.0058 & 0.4087 / 0.0057 & 0.4328 / 0.0043 & 0.4129 / 0.0043 & 0.4353 / 0.0044 \\
\multirow{3}{*}{InternLM3-8B} & 0.3 & 0.3287 / 0.0082 & 0.3293 / 0.0083 & 0.3298 / 0.0086 & 0.3237 / 0.0091 & 0.3268 / 0.0091 & 0.3314 / 0.0088 & 0.3301 / 0.0084 & 0.3232 / 0.0093 \\
 & 0.7 & 0.3270 / 0.0068 & 0.3242 / 0.0064 & 0.3186 / 0.0076 & 0.3266 / 0.0067 & 0.3225 / 0.0072 & 0.3231 / 0.0070 & 0.3184 / 0.0081 & 0.3176 / 0.0064 \\
 & 1.2 & 0.2627 / 0.0076 & 0.2636 / 0.0068 & 0.2670 / 0.0081 & 0.2722 / 0.0057 & 0.2675 / 0.0066 & 0.2618 / 0.0091 & 0.2640 / 0.0068 & 0.2665 / 0.0069 \\
\bottomrule
\end{tabular*}
\end{table*}

\clearpage

\begin{table*}[!t]\fontsize{7.2}{8.0}\selectfont
\centering
\refstepcounter{section}
\label{app:detailed-tables}
\setlength{\tabcolsep}{2pt}
\renewcommand{\arraystretch}{1.285}
\caption{\textbf{Appendix \thesection: Detailed Experimental Data Across All Models and Benchmarks.} Each centralized score averages ten fixed-configuration benchmark evaluations; each decentralized score uses CoEvalChain's collaborative aggregation with the same per-question repetition count. The ten scores per method and model--benchmark pair yield the outcome statistics in Table~\ref{tab:model_performance}.}
\begin{tabular*}{\textwidth}{@{\extracolsep{\fill}}c*{10}{c}@{}}
\toprule
\textbf{Model} & \multicolumn{2}{c}{\textbf{GSM8K}} & \multicolumn{2}{c}{\textbf{Livecodebench}} & \multicolumn{2}{c}{\textbf{Humaneval}} & \multicolumn{2}{c}{\textbf{GPQA-Diamond}} & \multicolumn{2}{c}{\textbf{MMLU}} \\
\cmidrule(lr){2-3} \cmidrule(lr){4-5} \cmidrule(lr){6-7} \cmidrule(lr){8-9} \cmidrule(lr){10-11} 
 & Decentralized & Centralized & Decentralized & Centralized & Decentralized & Centralized & Decentralized & Centralized & Decentralized & Centralized\\

\midrule
\multirow{10}{*}{\textbf{Qwen-2.5-7B}} 
& 91.25 & 91.2 & 18.66 & 19.9 & 83.78 & 84.76 & 34.69 & 36.4 & 72.56 & 74.5\\
& 91.15 & 91.2 & 19.73 & 18.3 & 83.72 & 84.15 & 36.01 & 35.9 & 72.54 & 74.6\\
& 91.01 & 91.3 & 18.99 & 15.3 & 84.28 & 83.54 & 35.00 & 34.8 & 72.38 & 74.5\\
& 90.92 & 91.5 & 19.08 & 19.9 & 84.33 & 85.37 & 34.14 & 35.9 & 72.62 & 74.6\\
& 91.29 & 91.2 & 18.57 & 17.3 & 83.62 & 85.37 & 34.10 & 36.4 & 72.60 & 74.4\\
& 90.95 & 91.5 & 18.51 & 16.8 & 84.35 & 85.98 & 33.97 & 31.3 & 72.44 & 74.2\\
& 91.22 & 91.6 & 18.92 & 19.1 & 84.11 & 80.49 & 36.00 & 34.3 & 72.34 & 74.4\\
& 91.00 & 91.5 & 19.15 & 18.6 & 83.78 & 81.1 & 33.84 & 32.3 & 72.43 & 74.8\\
& 91.08 & 91.3 & 19.30 & 20.2 & 84.22 & 81.71 & 33.45 & 33.8 & 72.37 & 74.4\\
& 91.10 & 91.2 & 19.06 & 19.1 & 84.43 & 83.54 & 34.79 & 37.4 & 72.34 & 74.4\\
\midrule
\multirow{10}{*}{\textbf{InternLM-3-8B}} 
& 85.71 & 84.8 & 16.36 & 18.1 & 75.37 & 72.0 & 36.51 & 36.4 & 71.61 & 71.5\\
& 85.44 & 84.5 & 16.62 & 15.3 & 74.97 & 70.1 & 36.93 & 41.9 & 71.55 & 71.8\\
& 85.84 & 84.3 & 16.40 & 14.5 & 74.44 & 70.1 & 36.14 & 32.8 & 72.87 & 72.1\\
& 85.34 & 85.0 & 16.55 & 15.3 & 75.15 & 73.0 & 35.94 & 35.8 & 71.34 & 71.5\\
& 85.44 & 84.3 & 16.49 & 14.5 & 75.40 & 71.3 & 36.84 & 37.4 & 71.73 & 72.1\\
& 85.70 & 84.6 & 16.81 & 14.5 & 74.96 & 68.9 & 36.38 & 38.9 & 72.06 & 71.7\\
& 85.39 & 85.1 & 16.60 & 17.6 & 74.82 & 70.7 & 37.45 & 39.9 & 72.28 & 72.2\\
& 85.35 & 84.4 & 16.88 & 15.3 & 75.37 & 73.2 & 37.10 & 34.8 & 71.93 & 72.7\\
& 85.47 & 85.1 & 16.38 & 16.0 & 75.87 & 71.3 & 37.18 & 39.3 & 71.55 & 71.3\\
& 85.42 & 84.4 & 16.25 & 14.5 & 75.30 & 72.4 & 37.32 & 34.4 & 72.10 & 71.8\\
\midrule
\multirow{10}{*}{\textbf{Llama-3-8B}} 
& 79.71 & 80.1 & 12.65 & 13.7 & 61.16 & 59.2 & 31.16 & 35.9 & 63.35 & 64.0\\
& 79.73 & 79.4 & 13.17 & 13.0 & 60.30 & 63.4 & 33.18 & 33.8 & 63.28 & 63.6\\
& 79.94 & 79.1 & 12.63 & 12.2 & 60.48 & 57.9 & 33.94 & 29.8 & 63.39 & 63.9\\
& 79.23 & 80.1 & 12.80 & 13.0 & 61.22 & 62.2 & 32.12 & 36.4 & 63.64 & 63.7\\
& 79.64 & 78.5 & 12.66 & 12.2 & 60.85 & 59.8 & 33.28 & 33.8 & 63.36 & 63.7\\
& 79.70 & 79.2 & 12.93 & 11.5 & 61.43 & 62.8 & 34.49 & 29.8 & 63.36 & 63.9\\
& 79.79 & 79.4 & 12.51 & 12.2 & 60.66 & 54.9 & 34.64 & 34.8 & 63.67 & 63.6\\
& 79.69 & 79.3 & 12.54 & 12.2 & 60.59 & 57.9 & 32.11 & 32.3 & 63.59 & 63.9\\
& 79.58 & 79.7 & 12.75 & 13.7 & 60.43 & 61.6 & 35.15 & 30.3 & 63.38 & 63.6\\
& 79.79 & 81.1 & 12.86 & 11.1 & 61.81 & 59.8 & 33.00 & 30.0 & 63.36 & 63.9\\
\midrule
\multirow{10}{*}{\textbf{Phi-4-Mini}} 
& 84.73 & 84.8 & 16.48 & 21.4 & 68.72 & 65.2 & 35.30 & 36.4 & 69.36 & 71.4\\
& 83.36 & 85.1 & 16.86 & 19.1 & 68.78 & 65.8 & 34.85 & 35.9 & 68.31 & 71.7\\
& 85.25 & 84.4 & 17.34 & 19.9 & 68.05 & 66.5 & 33.38 & 36.4 & 68.48 & 71.2\\
& 84.70 & 84.1 & 17.26 & 18.3 & 70.18 & 67.1 & 34.75 & 38.9 & 68.34 & 71.8\\
& 83.84 & 84.3 & 17.11 & 16.8 & 69.88 & 64.6 & 34.24 & 40.4 & 68.30 & 71.6\\
& 84.73 & 84.9 & 17.13 & 18.3 & 68.00 & 65.2 & 33.87 & 35.9 & 68.48 & 71.5\\
& 84.27 & 84.6 & 17.37 & 18.3 & 69.58 & 64.6 & 35.62 & 35.4 & 68.52 & 71.8\\
& 85.12 & 83.9 & 16.94 & 20.6 & 67.88 & 62.2 & 32.57 & 31.3 & 68.38 & 71.0\\
& 83.70 & 85.0 & 17.15 & 21.4 & 70.12 & 60.4 & 36.29 & 33.3 & 68.50 & 71.9\\
& 85.14 & 84.8 & 17.36 & 16.5 & 67.92 & 66.5 & 31.94 & 32.0 & 68.30 & 72.0\\
\midrule
\multirow{10}{*}{\textbf{Qwen-3-14B}} 
& 95.87 & 96.1 & 43.68 & 53.4 & 94.02 & 93.9 & 58.94 & 61.1 & 82.49 & 85.2\\
& 95.77 & 96.4 & 43.05 & 54.2 & 93.51 & 97.0 & 60.31 & 58.6 & 82.92 & 84.9\\
& 95.86 & 96.1 & 42.76 & 54.2 & 93.14 & 98.2 & 59.15 & 59.1 & 82.67 & 85.1\\
& 95.77 & 96.4 & 43.62 & 51.9 & 93.71 & 96.3 & 58.50 & 61.6 & 82.61 & 84.8\\
& 95.72 & 96.1 & 43.32 & 51.9 & 92.96 & 94.5 & 58.65 & 54.2 & 82.53 & 84.9\\
& 96.07 & 96.1 & 43.95 & 55.0 & 94.24 & 97.6 & 60.53 & 60.6 & 82.53 & 84.7\\
& 95.77 & 95.8 & 42.95 & 51.9 & 93.23 & 93.3 & 58.68 & 63.6 & 82.50 & 85.3\\
& 96.03 & 95.7 & 43.08 & 61.1 & 93.32 & 95.1 & 58.29 & 63.1 & 82.43 & 85.0\\
& 96.09 & 96.4 & 44.05 & 60.1 & 94.04 & 95.7 & 60.62 & 64.6 & 82.94 & 85.2\\
& 95.84 & 96.3 & 44.01 & 53.4 & 93.27 & 96.3 & 60.16 & 51.1 & 82.67 & 84.8\\

\bottomrule
\end{tabular*}
\label{tab:detailed_model_performance}
\end{table*}

\end{document}